\documentclass[10pt,twocolumn,letterpaper]{article}

\usepackage{cvpr}              

\usepackage{arydshln}
\usepackage{graphicx}
\usepackage{colortbl}
\usepackage{booktabs}
\usepackage{multirow} 
\usepackage{array}
\usepackage{colortbl}
\usepackage{pifont}    
\usepackage{bbding}    
\usepackage{placeins}  

\newcommand{\cmark}{\ding{51}\xspace}%
\newcommand{\xmarkg}{\textcolor{lightgray}{\ding{55}}\xspace}

\definecolor{cvprblue}{rgb}{0.21,0.49,0.74}
\usepackage[pagebackref,breaklinks,colorlinks,allcolors=cvprblue]{hyperref}

\definecolor{cvprblue}{rgb}{0.21,0.49,0.74}
\usepackage[pagebackref,breaklinks,colorlinks,allcolors=cvprblue]{hyperref}
\definecolor{tabblue}{HTML}{1f77b4}
\definecolor{taborange}{HTML}{ff7f0e}
\definecolor{tabgreen}{HTML}{2ca02c}
\definecolor{tabred}{HTML}{d62728}
\definecolor{tabpurple}{HTML}{9467bd}
\definecolor{tabpink}{HTML}{ff0080}
\usepackage[accsupp]{axessibility} 

\def\paperID{} 
\def\confName{CVPR}
\def\confYear{2026}

\title{MARIS: \underline{Mar}ine Open-Vocabulary \underline{I}nstance \underline{S}egmentation}

\author{
Bingyu Li\textsuperscript{1,2}\thanks{Work done during an internship at TeleAI.} \quad
Feiyu Wang\textsuperscript{3,2} \quad
Da Zhang\textsuperscript{4,2} \quad
Zhiyuan Zhao\textsuperscript{2} \quad
Junyu Gao\textsuperscript{2} \quad
Xuelong Li\textsuperscript{2}\thanks{Corresponding Author} \\
\textsuperscript{1}University of Science and Technology of China, China \\
\textsuperscript{2}Institute of Artificial Intelligence (TeleAI), China Telecom, China \\
\textsuperscript{3}Fudan University, China \\
\textsuperscript{4}Northwestern Polytechnical University, China \\
}

\begin{document}
\maketitle
\begin{abstract}
Most existing underwater instance segmentation approaches are constrained by close-vocabulary prediction, limiting their ability to recognize novel marine categories.
To support evaluation, we introduce \textbf{MARIS} (\underline{Mar}ine Open-Vocabulary \underline{I}nstance \underline{S}egmentation), the first large-scale fine-grained benchmark for underwater Open-Vocabulary (OV) Instance segmentation (UOVIS), featuring a limited set of seen categories and diverse unseen categories. 
Although OV instance segmentation has shown promise on natural images, our analysis reveals that transfer to underwater scenes suffers from severe visual degradation (e.g., color attenuation) and semantic misalignment caused by lack underwater class definitions.
To address these issues, we propose a unified framework with two complementary components. The Geometric Prior Enhancement Module (\textbf{GPEM}) leverages stable part-level and structural cues to maintain object consistency under degraded visual conditions. The Semantic Alignment Injection Mechanism (\textbf{SAIM}) enriches language embeddings with domain-specific priors, mitigating semantic ambiguity and improving recognition of unseen categories.
Experiments show that our framework consistently outperforms existing OV baselines both In-Domain and Cross-Domain setting on MARIS, establishing a strong foundation for future underwater perception research. The code is \href{https://github.com/LiBingyu01/MARIS}{\textcolor{tabpink}{Here}}\footnote{\textit{\textcolor{tabpink}{https://github.com/LiBingyu01/MARIS}}}.
\end{abstract}
   
\section{Introduction}
\label{sec:intro}
Instance segmentation in underwater imagery plays a crucial role in applications such as marine biodiversity monitoring, autonomous underwater vehicles, and environmental conservation \cite{li2025uwsam, hong2024watersam}. The goal of this task is to accurately localize and categorize marine objects with pixel-level instance masks. However, existing approaches heavily rely on dense pixel-wise annotations, which are extremely costly to obtain in underwater environments\cite{li2025u3m}. Furthermore, conventional models are limited by the restricted set of training categories, hindering their ability to generalize to unseen species or adapt to novel marine exploration scenarios\cite{ge2025underwater, abdullah2024caveseg}.

\begin{figure}
  \centering
  \includegraphics[width=0.9\linewidth]{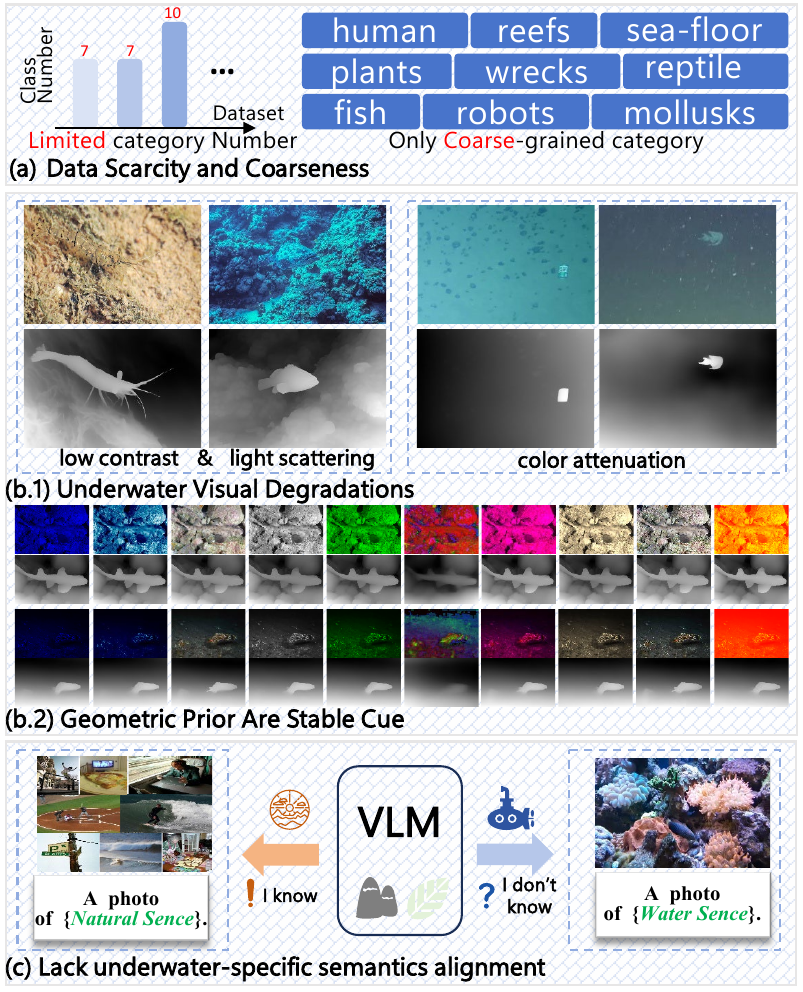}
  \vspace{-8pt}
  \caption{\textbf{The challenges of transferring OV instance segmentation to underwater scenarios} in terms of (a) datasets and (b-c) methods, which have motivated the contributions of this study.}
  \label{fig:fig_moti}
\vspace{-10pt}
\end{figure}

OV learning \cite{chen2025efficient, cho2024cat, yu2023convolutions} offers a promising solution by enabling models to recognize novel categories without exhaustive labeling or retraining. While OV segmentation models have demonstrated strong performance on terrestrial and natural images, their direct transfer to underwater imagery remains unexplored. 

We analyze the OV learning paradigm in the context of underwater scenarios and identify several key challenges. The first challenge is \textbf{data scarcity and coarse-grained annotations}: OV segmentation typically relies on large-scale\cite{niu2025eov}, diverse annotations\cite{ding2022open}, as illustrated in \cref{fig:fig_moti}(a) existing underwater datasets, such as UIIS10K \cite{li2025uwsam} and USIS10K \cite{lian2024diving}, provide labels for only less than 20 categories.
Moreover, many underwater organisms are crudely grouped into broad classes such as “fish” and “plants.” For instance, \texttt{Amphora} and \texttt{Blue Parrotfish} are just categorized as “fish,”. This coarse labeling severely restricts OV transfer. To overcome this limitation, we present the MARIS dataset, which introduces 158 fine-grained category labels with diverse instances, establishing the first benchmark for OV segmentation in underwater environments.

Even with sufficiently annotated data, transferring models to underwater imagery remains challenging due to the unique characteristics of underwater environments\cite{xue2025uvlm,truong2025autv}. Unlike terrestrial images, underwater images are captured through a medium(water) that induces significant visual degradations\footnote{\textit{color attenuation}, \textit{low contrast}, and \textit{light scattering}} in \cref{fig:fig_moti}(b.1). For instance, organisms whose body colors closely resemble the surrounding environment can become visually indistinguishable, and objects may become partially or fully occluded due to lighting conditions or water turbidity. In essence, such degradations render \textbf{visual appearance cues unstable} in underwater scenes.

On the other hand, despite these visual degradations, many underwater objects retain stable geometric properties that can serve as reliable cues. As shown in \cref{fig:fig_moti}(b.2), our preliminary visualization experiments demonstrate that although fish may lose distinctive color patterns, their body shapes and fin structures remain discernible. Likewise, coral colonies exhibit characteristic geometric growth patterns even when their surface textures are degraded. Motivated by this observation, we propose a \textbf{Geometric Prior Enhancement Module (GPEM)}, which exploits geometric priors to alleviate visual degradations in underwater imagery.

Beyond visual degradation, another distinct property of underwater imagery is \textbf{semantic ambiguity} caused by and insufficient language priors. As shown in \cref{fig:fig_moti}(c), current VLM, trained primarily on terrestrial data, fail to capture such fine-grained marine semantics. Motivated by this, we propose a \textbf{Semantic Alignment Injection Mechanism (SAIM)}, which integrates domain-specific knowledge via prompt augmentation and embedding enrichment. By guiding the model with enriched underwater semantics, SAIM mitigates category ambiguity and improves recognition of unseen species.  
Together, GPEM and SAIM function complementarily, addressing the core challenges of visual degradation and semantic ambiguity in underwater imagery from distinct yet synergistic perspectives.

Our contributions can be summarized as follows:
\begin{itemize}
    \item \textbf{New benchmark.} We introduce \textbf{MARIS}, the first large-scale fine-grained dataset for OV underwater instance segmentation, addressing the limitations of existing datasets with coarse-grained annotations.  
    \item \textbf{Novel framework.} We propose two complementary modules: \textbf{GPEM}, which leverages stable geometric priors to alleviate the impact of underwater visual degradations, and \textbf{SAIM}, which integrates domain-specific semantic knowledge to resolve ambiguity in marine category recognition.
    \item \textbf{Comprehensive evaluation.} Extensive experiments on MARIS demonstrate that our framework achieves state-of-the-art performance on underwater instance segmentation and shows strong generalization to unseen marine categories.
\end{itemize}

\begin{figure*}
  \centering
  \includegraphics[width=0.9\linewidth]{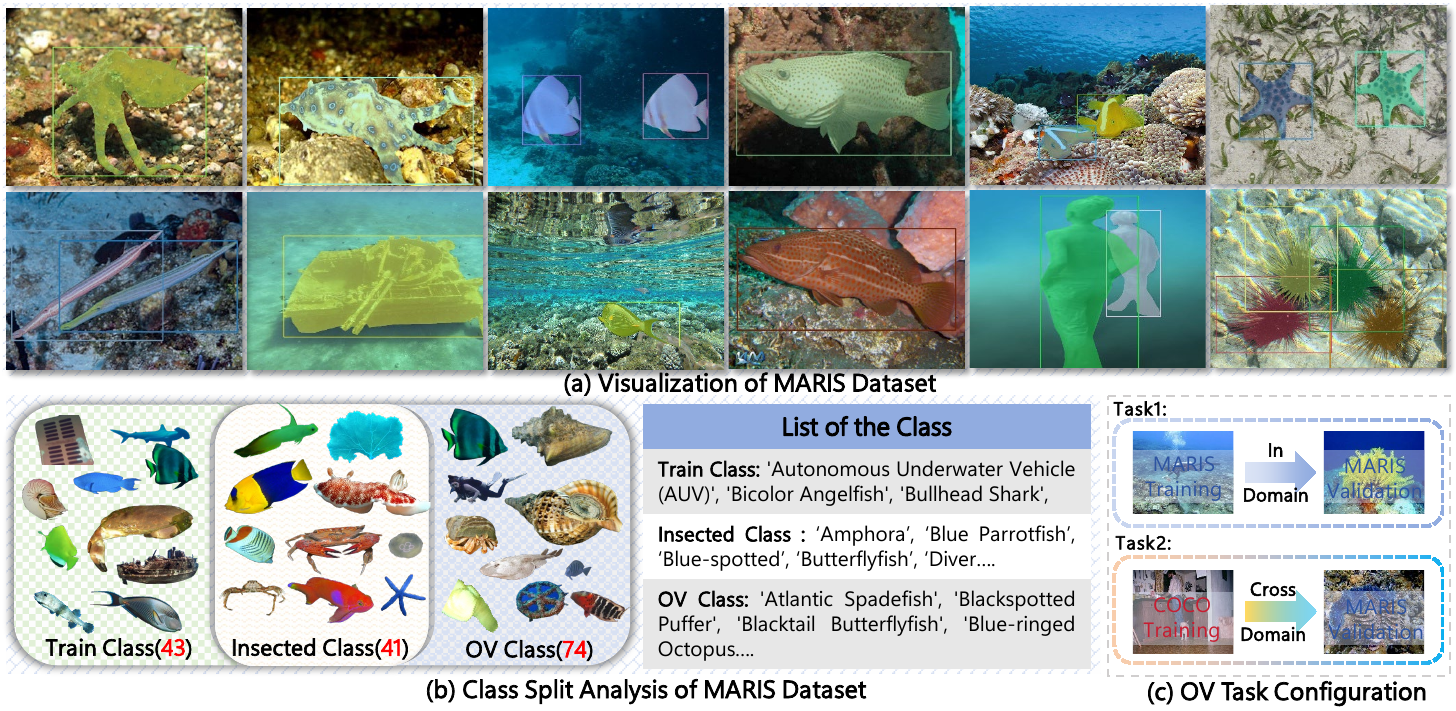}
  \vspace{-7pt}
  \caption{\textbf{Visualization and analysis of the MARIS dataset.} (a) Sample images from the MARIS dataset with object annotations. (b) Class split analysis, including Train Class, Insected Class, and OV Class. (c) Configuration of OV tasks, covering in-domain and cross-domain settings.}
  \label{fig:fig_dataset}
  \vspace{-7pt}
\end{figure*}

\section{Related Work}
\label{sec:rela_work}
\paragraph{Underwater Segmentation}
Underwater scene segmentation has been supported by several datasets. Early benchmarks such as SUIM \cite{islam2020semantic}, MAS3K \cite{fu2023masnet}, and DUT-USEG \cite{ma2021underwater} provided foundational data but were limited in category diversity or annotation quality. More recent efforts, including UIIS \cite{lian2023watermask}, UIIS10K \cite{li2025uwsam}, USIS10K \cite{lian2024diving}, and Seaclear \cite{djuravs2024dataset}, expanded scale and scope, while USIS16K \cite{hong2025usis16k} further introduced large-scale pixel-level salient instance masks with multi-level labels. Nonetheless, these datasets remain constrained for OV segmentation due to coarse taxonomies and limited category coverage.
Beyond data, underwater vision faces inherent challenges such as color attenuation, low contrast, and scattering. Traditional methods adapt general segmentation architectures with underwater-specific priors and enhancements \cite{yi2024coordinate, lian2023watermask, ge2025underwater, shi2024crackinst}. Representative models include UWSegFormer \cite{zuo2025improving}, UISS-Net \cite{he2024uiss}, and CaveSeg \cite{abdullah2024caveseg}.
Recently, Deep Learning \cite{zhang2023tdec, zhang2023cnmbi, zheng2024deep, du2026unsupervised} and Vision Foundation Models (VFMs)\cite{shen2025fine, li2026toward, liu2026palm, xu2025stare}, particularly SAM-based approaches \cite{li2025uwsam, lian2024diving, hong2024watersam}, have been adapted for underwater tasks. These developments highlight VFMs as a promising direction for robust, scalable segmentation in aquatic environments. 
Although underwater segmentation has progressed considerably, large-scale training for OV object segmentation remains unexplored. In this work, we take a step toward addressing this gap.

\paragraph{Open-Vocabulary Segmentation}
Open-Vocabulary Segmentation (OVS) seeks to segment image regions according to an open-world vocabulary, enabling generalization beyond pre-defined categories. Early works adapted vision-language models (VLMs)\cite{ma2026thinkingblueprintsassistingvisionlanguage, xie2026spatialqa, xie2025training, su2025large, wu2024detecting, zhang2025cross} such as CLIP \cite{radford2021learning} to pixel-level tasks. LSeg \cite{li2022language} employed pixel-wise contrastive learning for zero-shot segmentation, while proposal-based approaches, including MaskFormer \cite{cheng2021per} and ZSSeg \cite{xu2022simple}, generated class-agnostic masks for subsequent classification. FreeSeg \cite{qin2023freeseg} unified this paradigm with a one-shot framework maintaining consistent parameters across tasks. 
Later methods exploited dense features and improved efficiency.MaskCLIP \cite{dong2023maskclip} extracted patch-level features directly from CLIP, preserving vision-language alignment. SAN \cite{xu2023side} introduced side adapters into frozen CLIP backbones, while ODISE \cite{xu2023open} employed diffusion-based image-text embeddings for mask generation. Other one-stage methods \cite{yu2023convolutions, jiao2024collaborative}, extended the single-stage paradigm by introducing a matching loss to enforce better pixel–text alignment.
Recent work emphasized structural priors and cost aggregation. SCAN \cite{liu2024open} enhanced feature quality via self-supervised learning. Other methods such as CAT-Seg and ERR-Seg \cite{cho2024cat, chen2025efficient} transferred CLIP knowledge through cost aggregation without explicit mask categorization, reducing complexity \cite{li2025fgaseg, xie2024sed}. Other approaches, such as frequency-domain modules \cite{xu2024generalization} and adaptive fusion of SAM and CLIP outputs \cite{shan2024open}, further improved generalization and adaptability. 
In this paper, we make the first attempt to explore the OVS task in underwater scenarios and propose a novel model paradigm to adapt OVS models to the underwater domain.
\section{MARIS Benchmark}
\label{sec:dataset}
As a foundational step toward underwater OVS, we pioneer the construction of a dedicated benchmark, which incorporates precise evaluations.

\begin{figure*}
  \centering
  \includegraphics[width=1.0\linewidth]{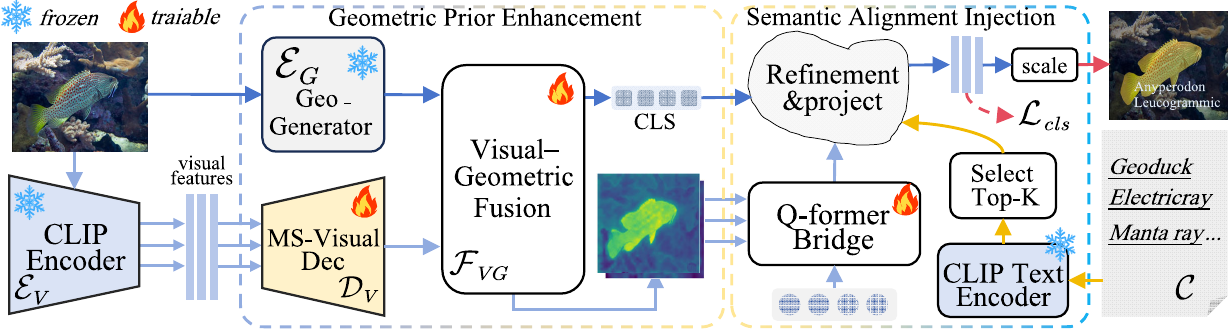}
  \vspace{-20pt}
  \caption{\textbf{Overall framework of the proposed Method.} The Geometric Prior Enhancement module strengthens structural representations via visual–geometric fusion and transformer-based query refinement. The Semantic Alignment Injection mechanism align category semantics with degraded underwater conditions.}
  \label{fig:fig_method}
  \vspace{-7pt}
\end{figure*}

\subsection{Data Collection and Annotation}
Our benchmark, \textbf{MARIS} (\underline{Mar}ine \underline{I}nstance \underline{S}egmentation), is developed to overcome the limitations of existing underwater segmentation benchmarks, which remain scarce and coarse-grained. Public datasets such as UIIS \cite{lian2023watermask} and USIS10K \cite{lian2024diving} contain fewer than 20 annotated categories and group diverse organisms into broad groups such as ``fish`` or ``plants`` class. Such coarse labeling restricts OV models from generalizing to unseen or fine-grained categories.  
To address this gap, MARIS (\cref{fig:fig_dataset}(a)) is curated from multiple complementary sources \cite{lian2024diving, lian2023watermask}, including several recently released underwater datasets \cite{islam2020semantic, hong2025usis16k, li2025uwsam}, which we systematically re-annotate and extend based on \cite{hong2025usis16k}. In total, MARIS comprises over 16K underwater images categorized into 9 super-classes and 158 fine-grained subclasses. Unlike prior benchmarks, our annotations explicitly distinguish detailed categories---for example, the ``fish’’ super-class is refined into 76 distinct species (see Appendix for details). This ensures coverage of diverse marine organisms, artificial objects, and natural substrates. We list some of the categories in \cref{fig:fig_dataset}(b).
All annotations are provided at the instance level with pixel-accurate masks, enabling detailed structural analysis. This fine-grained labeling not only enhances semantic richness but also establishes MARIS as \textit{the first benchmark} to support rigorous evaluation of OV instance segmentation in underwater environments.  

\subsection{Dataset Split and Experimental Settings}
The MARIS dataset contains 5,712 training images and 10,439 validation images. While the initial category ratio was designed as 1:2, the presence of multiple instances per image resulted in 84 training categories and 115 validation categories, with 41 overlapping between them. Consequently, shown in \cref{fig:fig_dataset}(b), the training set contains 43 exclusive classes, and the testing set contains 74 exclusive classes, more details are in the Appendix E-F.

\subsubsection{Task Configuration}
Based on this split, we define two experimental settings as illustrated in \cref{fig:fig_dataset} (c).
\textbf{In-domain.} For in-domain evaluation, models are trained on the MARIS training set and evaluated on the validation set.
\textbf{Cross-domain.} To further assess cross-domain generalization, we design a more challenging setting where models are trained on COCO\cite{lin2014microsoft} and evaluated on the MARIS validation set. Since COCO and MARIS share no category overlap, this configuration rigorously tests the ability of models to adapt from a generic dataset to the underwater domain. 

\section{Method}
\label{sec:method}

\subsection{Problem Definition}
Formally, given an input image $\mathbf{I}$ and a set of textual category descriptions $\mathcal{C} = \{c_1, c_2, \dots, c_n\}$, an OVIS model aims to produce a set of instance masks $\mathbf{M} = \{m_1, m_2, \dots, m_k\}$ and corresponding labels $\mathbf{Y} = \{y_1, y_2, \dots, y_k\}$, where each $y_i \in \mathcal{C}$ may represent categories that are unseen during training.

\subsection{Overall Architecture}

Given an input underwater image $\mathbf{I}$, the processing pipeline of MARIS can be expressed as:

\begin{equation}
\mathbf{F}_G = \mathcal{E}_G(\mathbf{I}), \quad 
\mathbf{F}_V = \mathcal{E}_V(\mathbf{I}),
\end{equation}
where $\mathbf{F}_G$ denotes the geometric prior features extracted by the frozen Geo-Generator, and $\mathbf{F}_V$ represents the visual features from the frozen CLIP visual encoder. The multi-scale visual decoder $\mathcal{D}_V$ processes $\mathbf{F}_V$ and fuses it with $\mathbf{F}_G$ via the visual-geometric fusion module $\mathcal{F}_{VG}$:

\begin{equation}
\mathbf{F}_{VG} = \mathcal{F}_{VG} \big( \mathcal{D}_V(\mathbf{F}_V), \mathbf{F}_G \big),
\end{equation}
producing the enhanced visual-geometric representation $\mathbf{F}_{VG}$ along with a global [CLS] token. The Semantic Alignment Injection Mechanism (SAIM) then refines these features with semantic embeddings $\mathbf{E}_{T}$ generated by the frozen CLIP text encoder:

\begin{equation}
(\mathbf{Y}_{\text{cls}}, \mathbf{M})= \text{SAIM}(\mathbf{F}_{VG}, \mathbf{E}_T).
\end{equation}

The refined feature representation $\mathbf{Y}_{\text{cls}}$ and $\mathbf{M}$ are used to jointly supervise the model through the classification loss $\mathcal{L}_{cls}$ and the mask loss $\mathcal{L}_{mask}$.

\subsection{Geometric Prior Enhancement Module}
The GPEM is designed for fuse multi-scale CLIP visual features with depth-derived geometric priors, producing enhanced representations that combine semantic context with structural information.

\paragraph{Multi-scale Visual \& Geometric  Generator}  
Given hierarchical features $\{\mathbf{F}_V^{(l)}\}_{l=1}^{L}$ extracted by the frozen CLIP encoder: $\{\mathbf{F}_V^{(l)}\}_{l=1}^{L} = \mathcal{E}_V(\mathbf{I})$,
we employ a multi-scale deformable attention module to refine local details and long-range dependencies. The outputs include enhanced features at each scale and an aggregated global visual representation $\mathbf{F}_m$:
\begin{equation}
\{\{\tilde{\mathbf{F}}_V^{(l)}\}_{l=1}^{L}, \mathbf{F}_m \} = \mathrm{MS\mbox{-}DeformAttn}\!\left(\{\mathbf{F}_V^{(l)}\}_{l=1}^{L}\right).
\end{equation}

To incorporate reliable structural cues, we use a frozen depth encoder \cite{yang2024depthv2, yang2024depth} to produce multi-scale geometric features $\{\mathbf{F}_G^{(l)}\}_{l=1}^{L}$ and a global depth token $\mathbf{g}_{\text{cls}}$:
\begin{equation}
\{\{\mathbf{F}_G^{(l)}\}_{l=1}^{L}, \mathbf{g}_{\text{cls}}\} = \mathcal{E}_G(\mathbf{I}).
\end{equation}

\paragraph{Visual–Geometric Feature Fusion $\mathcal{F}_{VG}$:}  
To integrate multi-scale visual and geometric representations, both modalities are first projected into a shared latent space:
\begin{equation}
\hat{\mathbf{F}}_V^{(l)} = W_V^{(l)} \tilde{\mathbf{F}}_V^{(l)}, \qquad
\hat{\mathbf{F}}_G^{(l)} = W_G^{(l)} \mathbf{F}_G^{(l)}.
\end{equation}

An adaptive weight is then computed for each scale:
\begin{equation}
\alpha^{(l)} = \sigma\!\left(W_{\alpha}^{(l)}[\hat{\mathbf{F}}_V^{(l)} \,\|\, \hat{\mathbf{F}}_G^{(l)}]\right),
\end{equation}
and the fused feature is obtained as:
\begin{equation}
\mathbf{F}_{VG}^{(l)} = \mathrm{MLP}\Big(\hat{\mathbf{F}}_V^{(l)} + \alpha^{(l)} \odot \hat{\mathbf{F}}_G^{(l)}\Big),
\end{equation}
where $\sigma$ denotes the sigmoid function, $\|\,$ indicates concatenation, and $\odot$ is element-wise multiplication. This formulation allows multi-scale geometric cues to be adaptively injected, ensuring that structural depth information complements fine-grained visual features effectively.

\paragraph{Geometry-based Visual \& Semantic Bridge}  
To extract effective visual representations and bridge them with semantic information, we employ a lightweight Q-Former (a $N$-layer transformer encoder always used in VLM \cite{li2023blip, dai2023instructblip} to bridge visual and semantic features). The fused geometric-visual features $\mathbf{F}_{VG}^{(l)}$ are processed by the Q-Former to update the query embeddings $\mathbf{Q} \in \mathbb{R}^{N_Q \times C}$, and the final geometry-informed queries are obtained by aggregating outputs across all scales.

\begin{table*}[ht]
\centering
\small
\caption{\textbf{Comparison of in-domain open-vocabulary segmentation performance} across different methods and backbones. Our method consistently outperforms previous approaches on both ConvNext-B and ConvNext-L backbones. Rows with gray background highlight our method and its improvement over the second-best approach.}
\vspace{-1em}
\label{tab:main_table_id}
\resizebox{0.95\linewidth}{!}{%
\begin{tabular}{l l l r r r r r r r r r}
\toprule
\multirow{2}{*}{Method} & \multirow{2}{*}{Publication} & \multirow{2}{*}{Backbone} 
& \multicolumn{3}{c}{Intersection Class} 
& \multicolumn{3}{c}{Open-Vocabulary Class} 
& \multicolumn{3}{c}{Overall Class} \\
\cmidrule(lr){4-6} \cmidrule(lr){7-9} \cmidrule(lr){10-12}
 &  &  & mAP & AP$_{50}$ & AP$_{75}$ & mAP & AP$_{50}$ & AP$_{75}$ & mAP & AP$_{50}$ & AP$_{75}$ \\
\midrule
OVSeg\cite{liang2023open}        & CVPR'23 & ViT-B      & 37.52 & 48.51 & 43.26 & 27.21 & 33.65 & 30.38 & 30.95 & 39.02 & 35.47 \\
ODISE\cite{xu2023open}           & CVPR'23 & ViT-B      & 41.89 & 50.74 & 46.83 & 30.26 & 35.68 & 32.54 & 34.71 & 41.56 & 38.12 \\
SAN\cite{xu2023side}             & CVPR'23 & ViT-B      & 43.26 & 52.18 & 48.05 & 31.57 & 37.09 & 34.02 & 36.05 & 43.06 & 39.26 \\
FCCLIP\cite{yu2023convolutions}  & NeurIPS'23 & ConvNext-B & 47.78 & 57.22 & 52.44 & 34.53 & 39.84 & 37.15 & 39.26 & 46.03 & 42.60 \\
MAFT+\cite{jiao2024collaborative} & ECCV'24 & ConvNext-B & 48.15 & 58.26 & 54.57 & 35.72 & 40.67 & 38.88 & 40.08 & 47.16 & 43.33 \\
EOVSeg\cite{niu2025eov}         & AAAI'25 & ConvNext-B & 37.98 & 48.95 & 41.55 & 27.48 & 33.89 & 29.56 & 31.22 & 39.26 & 33.83 \\
\rowcolor{gray!10} 
Our Method  & --- & ConvNext-B & \textbf{52.68} & \textbf{61.56} & \textbf{57.33} & \textbf{39.77} & \textbf{45.78} & \textbf{42.68} & \textbf{44.37} & \textbf{51.41} & \textbf{47.90} \\
\rowcolor{gray!10} 
\textit{Ours vs 2nd} & --- & --- & \textcolor{blue}{↑4.53} & \textcolor{blue}{↑3.30} & \textcolor{blue}{↑2.76} & \textcolor{blue}{↑4.05} & \textcolor{blue}{↑5.11} & \textcolor{blue}{↑3.80} & \textcolor{blue}{↑4.29} & \textcolor{blue}{↑4.25} & \textcolor{blue}{↑4.57} \\
\midrule
OVSeg\cite{liang2023open}        & CVPR'23 & ViT-B      & 48.96 & 57.92 & 53.64 & 44.63 & 51.89 & 48.25 & 46.41 & 54.23 & 50.36 \\
ODISE\cite{xu2023open}           & CVPR'23 & ViT-B      & 49.32 & 58.75 & 54.26 & 45.18 & 52.64 & 48.93 & 46.95 & 55.02 & 51.07 \\
SAN\cite{xu2023side}             & CVPR'23 & ViT-B      & 50.17 & 59.63 & 55.08 & 46.05 & 53.47 & 49.76 & 47.78 & 55.86 & 51.92 \\
FCCLIP\cite{yu2023convolutions}  & NeurIPS'23 & ConvNext-L & 54.29 & 63.33 & 58.37 & 50.99 & 58.66 & 54.57 & 52.17 & 60.33 & 55.92 \\
MAFT+\cite{jiao2024collaborative} & ECCV'24 & ConvNext-L & 55.32 & 64.24 & 59.42 & 51.54 & 59.44 & 55.74 & 53.41 & 61.36 & 58.88 \\
EOVSeg\cite{niu2025eov}         & AAAI'25 & ConvNext-L & 51.72 & 63.16 & 55.57 & 48.32 & 57.26 & 51.53 & 49.53 & 59.36 & 53.04 \\
\rowcolor{gray!10} 
Our Method  & --- & ConvNext-L & \textbf{61.55} & \textbf{71.02} & \textbf{66.04} & \textbf{54.02} & \textbf{61.54} & \textbf{57.44} & \textbf{56.71} & \textbf{64.92} & \textbf{60.51} \\
\rowcolor{gray!10} 
\textit{Ours vs 2nd} & --- & --- & \textcolor{blue}{↑6.23} & \textcolor{blue}{↑6.78} & \textcolor{blue}{↑6.62} & \textcolor{blue}{↑2.48} & \textcolor{blue}{↑2.10} & \textcolor{blue}{↑1.70} & \textcolor{blue}{↑3.30} & \textcolor{blue}{↑3.56} & \textcolor{blue}{↑1.63} \\
\bottomrule
\end{tabular}%
}
\vspace{-10pt}
\end{table*}

\begin{table}[ht]
\centering
\small
\caption{\textbf{Cross-domain open-vocabulary segmentation results.} All models are trained on COCO and evaluated on the MARIS validation set. Rows with gray background highlight our method and its improvement over the second-best approach.}
\label{tab:main_table_cd}
\resizebox{0.95\linewidth}{!}{%
\begin{tabular}{lll r r r}
\toprule
\multirow{2}{*}{Method} & \multirow{2}{*}{Publication} & \multirow{2}{*}{Backbone} 
& \multicolumn{3}{c}{Overall Class} \\
\cmidrule(lr){4-6}
 &  &  & mAP & AP$_{50}$ & AP$_{75}$ \\
\midrule
OVSeg\cite{liang2023open}        & CVPR'23 & ViT-B      & 18.95 & 24.30 & 19.82 \\
ODISE\cite{xu2023open}           & CVPR'23 & ViT-B      & 18.51 & 23.86 & 19.40 \\
SAN\cite{xu2023side}             & CVPR'23 & ViT-B      & 19.18 & 24.63 & 20.05 \\
FCCLIP\cite{yu2023convolutions}  & NeurIPS'23 & ConvNeXt-B & 29.79 & 36.12 & 33.50 \\
MAFT+\cite{jiao2024collaborative} & ECCV'24 & ConvNeXt-B & 30.05 & 36.57 & 34.11 \\
EOVSeg\cite{niu2025eov}         & AAAI'25 & ConvNeXt-B & 18.90 & 25.91 & 21.19 \\
\rowcolor{gray!10} 
Our Method                       & ---     & ConvNeXt-B & \textbf{32.62} & \textbf{39.60} & \textbf{36.65} \\
\rowcolor{gray!10} 
\textit{Ours vs 2nd}            & ---     & ---        & \textcolor{blue}{↑2.57} & \textcolor{blue}{↑3.03} & \textcolor{blue}{↑2.54} \\
\midrule
OVSeg\cite{liang2023open}        & CVPR'23 & ViT-B & 30.65 & 40.78 & 37.90 \\
ODISE\cite{xu2023open}           & CVPR'23 & ViT-B & 32.82 & 41.95 & 37.01 \\
SAN\cite{xu2023side}             & CVPR'23 & ViT-B & 34.05 & 42.20 & 38.26 \\
FCCLIP\cite{yu2023convolutions}  & NeurIPS'23 & ConvNeXt-L & 39.46 & 46.39 & 43.62 \\
MAFT+\cite{jiao2024collaborative} & ECCV'24 & ConvNeXt-L & 40.27 & 47.89 & 45.72 \\
EOVSeg\cite{niu2025eov}         & AAAI'25 & ConvNeXt-L & 35.90 & 45.33 & 40.11 \\
\rowcolor{gray!10} 
Our Method                       & ---     & ConvNeXt-L & \textbf{46.18} & \textbf{54.34} & \textbf{51.11} \\
\rowcolor{gray!10} 
\textit{Ours vs 2nd}            & ---     & ---        & \textcolor{blue}{↑5.91} & \textcolor{blue}{↑6.45} & \textcolor{blue}{↑5.39} \\
\bottomrule
\end{tabular}%
}
\vspace{-16pt}
\end{table}

\subsection{Semantic Alignment Injection Mechanism}

We design the \textbf{Semantic Alignment Injection Mechanism (SAIM)} from two complementary perspectives: (1) introducing underwater-aware textual prompts and adaptive template selection, and (2) incorporating geometry-based global priors to enrich category representations.

\paragraph{Adaptation to Underwater Scenes}  
Generic language prompts in VLMs often fail to capture underwater-specific semantics, where degradations such as scattering, low contrast, and color attenuation distort object appearance \cite{radford2021learning, li2022blip}. To address this, we introduce \textbf{underwater prompts} as environment-aware priors into the text encoder. These prompts encode five complementary aspects of underwater scenes: (\textit{i}) environmental context, (\textit{ii}) water medium and visibility, (\textit{iii}) illumination and perception, (\textit{iv}) depth cues, and (\textit{v}) scene interactions, producing refined text embeddings that are consistent with underwater visual features.

Nevertheless, upon closer examination, we found that not all templates contribute equally; some may even introduce noise under degraded conditions. For example, in low-light scenarios, certain images can be effectively matched with prompts such as $\texttt{a <class> in low visibility conditions}$, yet such matches tend to be diluted when averaged with other less relevant prompts. 
To adaptively select the most reliable templates, we compute the similarity between visual features and all textual templates for each category. We rank the templates according to the average similarity across spatial positions and select the top-$N$ templates with the highest scores (detailed in Appendix D).

\paragraph{Category Discrimination}  
We fuse the global depth token $\mathbf{g}_{\text{cls}}$ with the aggregated mask features $\mathbf{F}_m$ to obtain enhanced representations $\mathbf{F}_f$. The compact pooled feature $\mathbf{F}_c = \mathrm{Pool}(\mathbf{F}_f)$ is first combined with the adapted text embeddings $\mathbf{E}_{T}$ to produce the classification predictions:
\begin{equation}
\mathbf{Y}_{\text{cls}} = \mathbf{F}_c \odot \hat{\mathbf{E}} \in \mathbb{R}^{Q \times C}.
\end{equation}
Meanwhile, the global depth token $\mathbf{g}_{\text{cls}}$ is fused with the aggregated mask features $\mathbf{F}_m$ to guide the query embeddings $\mathbf{Q}$ and produce the mask: $\mathbf{M}\in \mathbb{R}^{Q \times H \times W}$.

\subsection{Training}
During training, the model is optimized with a classification loss $\mathcal{L}_{\text{cls}}$,
\begin{equation}
\mathcal{L}_{\text{cls}} = \text{CrossEntropy}(\mathbf{Y}_{\text{cls}}, \mathbf{Y}_{\text{gt}}).
\end{equation}
implemented as a binary cross-entropy between the predicted and ground-truth categories, and a mask loss $\mathcal{L}_{\text{mask}}$, 

\begin{equation}
\mathcal{L}_{\text{mask}} = \text{DiceLoss}(\mathbf{M}, \mathbf{M}_{\text{gt}}) + \text{BCE}(\mathbf{M}, \mathbf{M}_{\text{gt}}),
\end{equation}
following the same formulation as MaskFormer\cite{yu2023convolutions} to supervise the predicted instance masks. Both losses are combined to guide the model toward accurate category recognition and precise spatial segmentation.

\section{Experiments And Results}
\label{sec:experiments_results}
\subsection{Experimental Details}
All experiments are conducted on four NVIDIA RTX 4090 GPUs (24GB memory) with the batch size of 16. We evaluate two experimental settings (in- and cross-domian) to comprehensively assess the proposed approach. The reproduction of comparative methods is detailed in Appendix B.

\subsection{Main Experiments}
\paragraph{Experiments for In-Domain Task}
Table~\ref{tab:main_table_id} reports results on both intersection and OV categories. MARIS consistently outperforms all competing methods under different backbones. With ConvNeXt-B, MARIS achieves 52.68 mAP on intersection classes and 39.77 mAP on OV classes, surpassing the strongest baseline by over 4 points. The improvement is further amplified with ConvNeXt-L, where MARIS reaches 61.55 mAP and 54.02 mAP on intersection and OV categories, respectively. Overall, MARIS delivers the best results across all metrics, with particularly notable gains under AP$_{75}$, indicating more accurate and robust mask predictions. These results demonstrate that our method effectively \textit{enhances category discrimination and generalization}, leading to superior performance in underwater OV segmentation.

\paragraph{Experiments for Cross-Domain Task}
Table~\ref{tab:main_table_cd} reports the results of cross-domain OVS, where models are trained on COCO and evaluated on the MARIS validation set. As expected, transferring models across domains leads to a clear performance drop, reflecting the large domain gap between terrestrial and underwater imagery. Methods such as MAFT+ and FCCLIP demonstrate relatively strong generalization, achieving around $30\%$ mAP with ConvNeXt-B backbones. However, EOVSeg struggles significantly, indicating that techniques relying heavily on domain-specific cues may fail in cross-domain scenarios. In contrast, our proposed MARIS framework achieves the best performance across both ConvNeXt-B and ConvNeXt-L backbones, surpassing previous methods by a consistent margin. In particular, MARIS improves the overall mAP from $30.05$ to $32.62$ with ConvNeXt-B and from $40.27$ to $46.18$ with ConvNeXt-L, highlighting its \textit{effectiveness in handling the severe visual degradations and semantic discrepancies of underwater environments}.
\subsection{Ablation Experiments}
\paragraph{Ablation Study of GPEM and SAIM}
Table \ref{tab:ablation_gpem_saim} reports the impact of GPEM and SAIM on segmentation performance. The baseline without either module achieves the lowest scores. Incorporating  improves Intersection Class metrics, while SAIM mainly benefits intersection Class AP$_{50}$ and Overall Class mAP. Notably, the integration of GPEM or SAIM particularly strengthens the model’s ability to generalize to OV classes. Combining both modules leads to the best results, with intersection Class mAP of 61.55\% and OV Class mAP of 54.02\%, demonstrating their complementary effects for enhancing both intersection and OV segmentation.
\begin{table}[ht]
\centering
\small
\caption{\textbf{Ablation study on the effectiveness of GPEM and SAIM components.} All experiments use large backbones for both $\mathcal{E}_{G}$ and $\mathcal{E}_{V}$. Rows with gray background indicate the combination of both components, achieving the best performance.}
\label{tab:ablation_gpem_saim}
\resizebox{\linewidth}{!}{%
\begin{tabular}{c c r r r r r r r r r}
\toprule
\multirow{2}{*}{GPEM} & \multirow{2}{*}{SAIM} & \multicolumn{3}{c}{Intersection Class} & \multicolumn{3}{c}{Open-Vocabulary Class} & \multicolumn{3}{c}{Overall Class} \\
\cmidrule(lr){3-5} \cmidrule(lr){6-8} \cmidrule(lr){9-11}
& & mAP & AP$_{50}$ & AP$_{75}$ & mAP & AP$_{50}$ & AP$_{75}$ & mAP & AP$_{50}$ & AP$_{75}$ \\
\midrule
\xmarkg & \xmarkg & 54.29 & 63.33 & 58.37 & 50.99 & 58.66 & 54.57 & 52.17 & 60.33 & 55.92 \\
\cmark & \xmarkg & 60.05 & 68.62 & 64.61 & 52.19 & 58.63 & 56.05 & 54.99 & 62.19 & 59.10 \\
\xmarkg & \cmark & 60.88 & 70.07 & 64.84 & 52.16 & 58.89 & 55.59 & 55.27 & 62.88 & 58.89 \\
\rowcolor{gray!10} \cmark & \cmark & \textbf{61.55} & \textbf{71.02} & \textbf{66.04} & \textbf{54.02} & \textbf{61.54} & \textbf{57.44} & \textbf{56.71} & \textbf{64.92} & \textbf{60.51} \\
\bottomrule
\end{tabular}%
}
\vspace{-6pt}
\end{table}

\begin{figure}[ht]
\centering
\includegraphics[width=\linewidth]{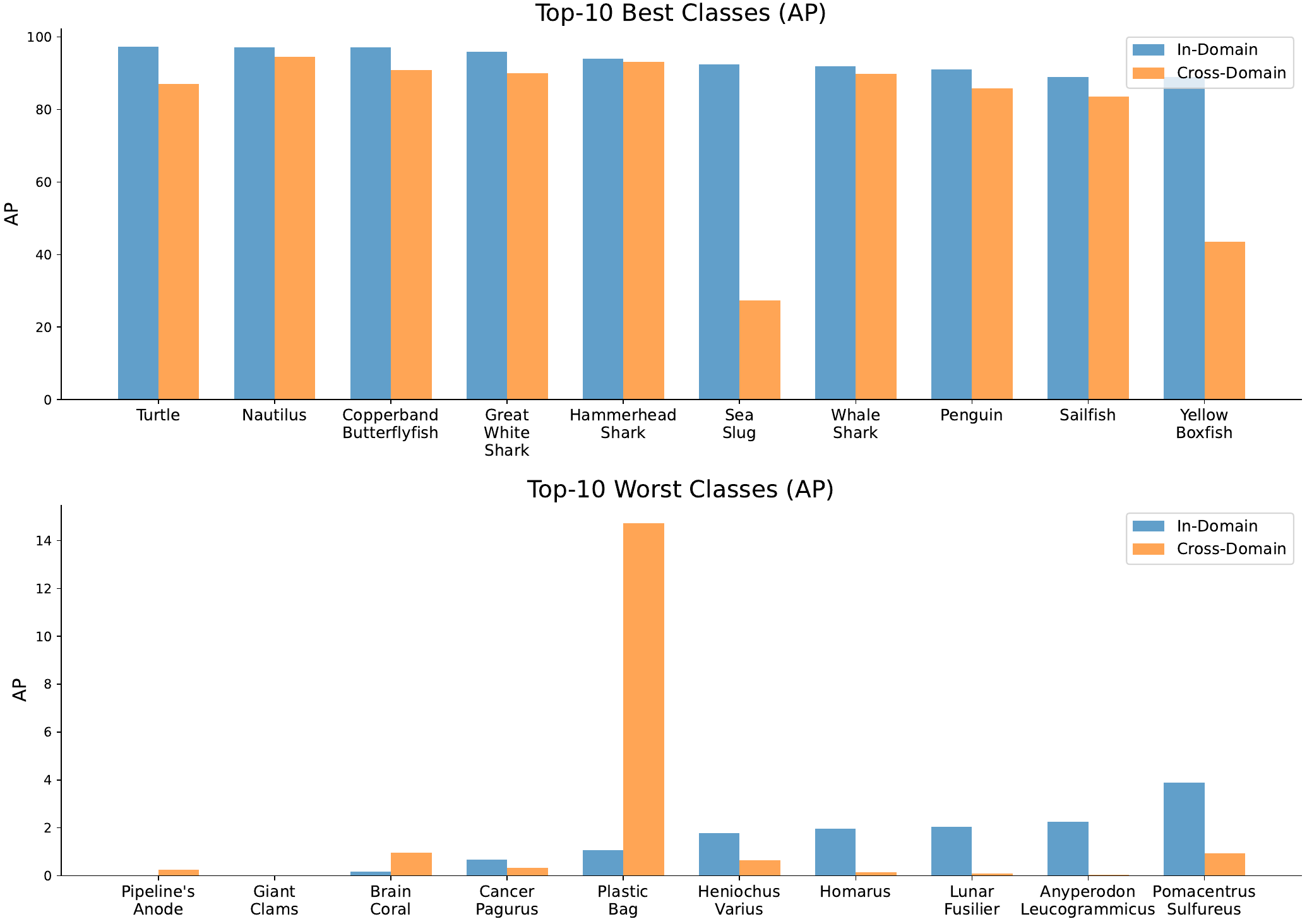}
\vspace{-20pt}
\caption{\textbf{Top-10 Best and Worst Classes:} Comparison of in-domain and cross-domain AP, illustrating performance drops and gains with geometric-enhanced fusion.}
\label{fig:pic_top10_per_class}
\vspace{-10pt}
\end{figure}

\paragraph{Effectiveness of Underwater Prompts and Template Selection}
Table \ref{tab:uw_prompt_ablation} evaluates different underwater prompt strategies. Adding underwater prompts (UW) already improves all metrics compared to using no prompts. Further, template selection consistently boosts performance. Notably, mixed selection strategy not only enhances general segmentation accuracy but also strengthens OV class performance, demonstrating its effectiveness for handling diverse underwater scenes.
\begin{table}[ht]
\centering
\small
\caption{\textbf{Ablation study on prompt strategies.} All experiments use base $\mathcal{E}_{G}$ and $\mathcal{E}_{V}$ models. The gray row highlights our final selection strategy. Bold values indicate the best results per column.}
\label{tab:uw_prompt_ablation}
\resizebox{\linewidth}{!}{%
\begin{tabular}{l r r r r r r r r r}
\toprule
Method & \multicolumn{3}{c}{Intersection Class} & \multicolumn{3}{c}{Open-Vocabulary Class} & \multicolumn{3}{c}{Overall Class} \\
\cmidrule(lr){2-4} \cmidrule(lr){5-7} \cmidrule(lr){8-10}
& mAP & AP$_{50}$ & AP$_{75}$ & mAP & AP$_{50}$ & AP$_{75}$ & mAP & AP$_{50}$ & AP$_{75}$ \\
\midrule
Template       & 51.92 & 60.74 & 56.31 & 37.92 & 42.82 & 40.60 & 42.91 & 49.21 & 46.20 \\
UWTemplate     & \textbf{53.99} & \textbf{62.92} & 58.10 & 38.29 & 43.88 & 40.97 & 43.89 & 50.67 & 47.08 \\
\rowcolor{gray!10} Selection & 53.80 & 62.35 & \textbf{59.04} & \textbf{39.40} & \textbf{44.99} & \textbf{42.35} & \textbf{44.54} & \textbf{51.17} & \textbf{48.30} \\
\bottomrule
\end{tabular}%
}
\vspace{-6pt}
\end{table}

\paragraph{Ablation Experiments of $\mathcal{E}_{G}$ size}
Table \ref{tab:ablation_eg_ev} shows that larger $\mathcal{E}_{G}$ (vitl) with Convnext-L yields the best in-domain results, while vitb consistently outperforms in cross-domain settings. This indicates that vitl benefits from higher capacity under matched distributions, but vitb strikes a better balance between capacity and generalization, reducing overfitting to in-domain patterns.

\paragraph{Ablation Experiments of Different feature fusion method}
Table~\ref{tab:5_MainAblation_Fvg} presents the ablation study on the proposed GPEM and SAIM. Without either component, the baseline achieves 52.17\% mAP overall. Introducing GPEM brings a clear improvement, raising the overall mAP to 54.99\%, which demonstrates its effectiveness in injecting global prompts to reduce domain discrepancies.
\begin{table}[ht]
  \centering
  \small
  \caption{\textbf{Ablation study on the Different $\mathcal{E}_{G}$ and $\mathcal{E}_{V}$ size.}}
\label{tab:ablation_eg_ev}
  \resizebox{\linewidth}{!}{
  \begin{tabular}{cccccccc}
    \toprule
    \multirow{2}{*}{$\mathcal{E}_{G}$} & \multirow{2}{*}{$\mathcal{E}_{V}$} & \multicolumn{3}{c}{in-Domain} & \multicolumn{3}{c}{Cross-Domain} \\
    \cmidrule(lr){3-5} \cmidrule(lr){6-8}
    & & $\text{mAP}$ & $\text{AP}_{50}$ & $\text{AP}_{75}$ & $\text{mAP}$ & $\text{AP}_{50}$ & $\text{AP}_{75}$ \\
    \midrule
    vits & ConvNext-B & 42.36 & 48.83 & 45.64 & 30.82 & 37.62 & 34.93 \\
    vitb & ConvNext-B & \cellcolor{gray!10}\textbf{44.54} & \cellcolor{gray!10}51.17 & \cellcolor{gray!10}\textbf{48.30} & \cellcolor{gray!10}\textbf{32.62} & \cellcolor{gray!10}\textbf{39.60} & \cellcolor{gray!10}\textbf{36.65} \\
    vitl & ConvNext-B & 44.37 & \textbf{51.41} & 47.90 & 32.07 & 38.55 & 35.73 \\
    \midrule
    vits & Convnext-L & 54.22 & 62.27 & 57.81 & 45.75 & 54.10 & 50.40 \\
    vitb & Convnext-L & 55.22 & 63.37 & 59.32 & \cellcolor{gray!10}\textbf{46.18} & \cellcolor{gray!10}\textbf{54.34} & \cellcolor{gray!10}\textbf{51.11} \\
    vitl & Convnext-L & \cellcolor{gray!10}\textbf{56.71} & \cellcolor{gray!10}\textbf{64.92} & \cellcolor{gray!10}\textbf{60.51} & 43.70 & 51.18 & 47.98 \\
    \bottomrule
  \end{tabular}
  }
\vspace{-10pt}
\end{table}
\begin{table}[ht]
\centering
\small
\caption{\textbf{Performance and efficiency comparison of different fusion methods.} We report overall-class metrics along with GFLOPS and model size. Rows with gray background indicate our proposed fusion method.}
\label{tab:5_MainAblation_Fvg}
\resizebox{0.95\linewidth}{!}{%
\begin{tabular}{l r r r r r}
\toprule
Method & mAP & AP$_{50}$ & AP$_{75}$ & GFLOPS & Params (M) \\
\midrule
MLP         & 43.87 & 50.73 & 47.36 & 364G & 21.72 \\
add         & 43.52 & 50.54 & 46.81 & 362G & 20.94 \\
\rowcolor{gray!10} alphafusion & \textbf{44.54} & \textbf{51.17} & \textbf{48.30} & 365G & 22.51 \\
\bottomrule
\end{tabular}%
}
\vspace{-6pt}
\end{table}

\begin{figure}[ht]
\centering
\includegraphics[width=\linewidth]{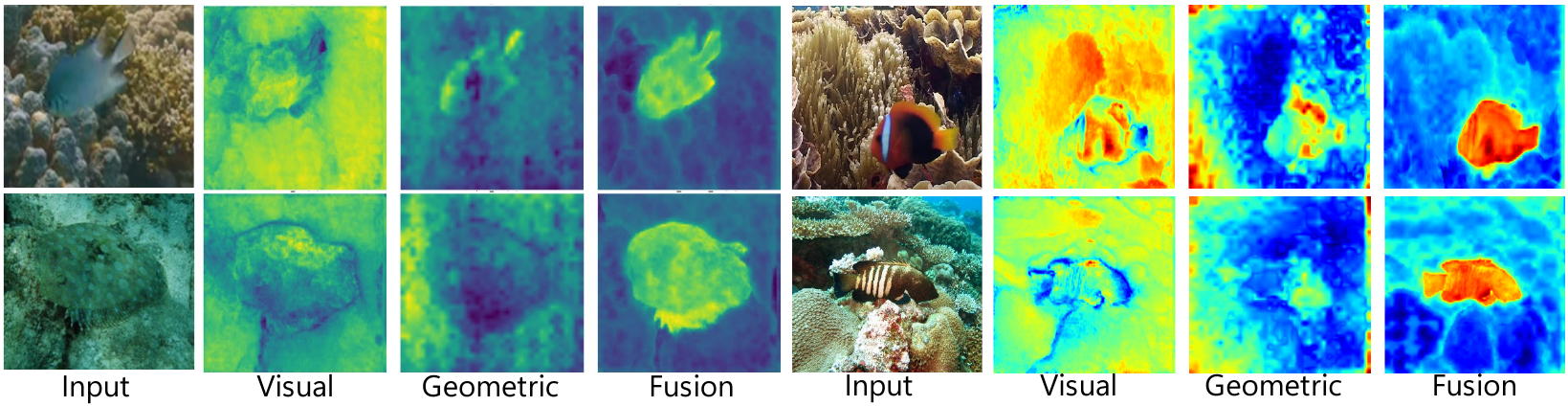}
\vspace{-2em}
\caption{\textbf{Qualitative Results} of visual information, geometric information, and their geometric-enhanced fusion, demonstrating clear improvements (viridis on the left and jet on the right).}
\label{fig:pic_vis_feats}
\vspace{-1em}
\end{figure}

\begin{figure*}[ht]
\centering
\includegraphics[width=0.87\linewidth]{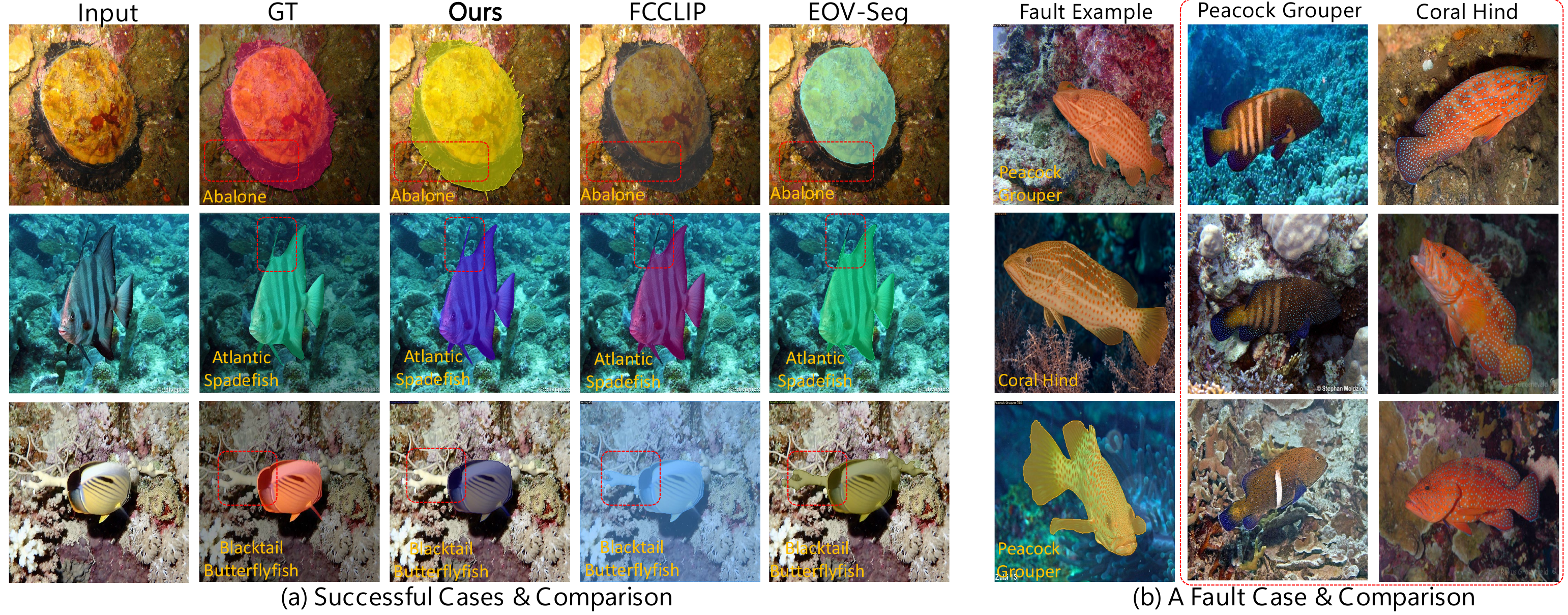}
\vspace{-10pt}
    \caption{
        \textbf{(a)} Successful cases and comparisons of our method with other approaches. 
        \textbf{(b)} A fault case where the model misclassifies \texttt{Anyperodon Leucogrammicus} as \texttt{Peacock Grouper} or \texttt{Coral Hind}. 
        Visually, these species share similarities, which likely leads to confusion in the model's prediction.
    }
\label{fig:pic_vis_seg}
\vspace{-10pt}
\end{figure*}

\subsection{Per-Class Performance Analysis}
The \cref{fig:pic_top10_per_class} highlights the top-10 and bottom-10 classes in terms of AP (More in Appendix J). Overall, high-frequency and visually distinctive categories (e.g., \textit{Shark}, \textit{Turtle}, \textit{Dolphin}) achieve consistently high AP across settings, indicating strong generalization. In contrast, rare or visually ambiguous categories (e.g., \textit{Sponges}, \textit{Anemonefish variants}, \textit{Small invertebrates}) exhibit large performance gaps, reflecting the challenges of recognition in underwater scenes.

\subsection{Cross-Domain and In-Domain Analysis}
\paragraph{Overall Performance Degradation in Cross-Domain}
In general, cross-domain performance is lower than in-domain, confirming the effectiveness of domain-specific knowledge. This suggests that incorporating more marine knowledge could further improve cross-domain generalization. On the other hand, it also indicates that our model, trained on natural scenes, can achieve effective cross-domain recognition.
\paragraph{Per-Class Failure Case Analysis}
We observed several failure cases where AP approaches zero, mostly corresponding to highly specialized species. Small fish such as \textit{Lunar Fusilier} and \textit{Pomacentrus Leucogrammicus} are not well captured by existing VLMs, likely due to insufficient semantic encoding. These cases highlight the challenges in cross-domain generalization caused by missing semantic alignment.
\paragraph{Cross-Domain Outperforming In-Domain}
Interestingly, \textit{Plastic Bag} achieves higher AP in cross-domain evaluation, likely because this object also appears in natural scenes (e.g., COCO dataset). This demonstrates that our model can effectively recognize objects in a new domain if they have been seen during training.

\subsection{Analysis of Inference Efficiency and Model Complexity}
As shown in Table~\ref{tab:FPS_FLOPS_Params}, our method consistently achieves higher in-domain mAP across different backbones. Despite the performance gains, it maintains lower GFLOPS and significantly fewer trainable parameters compared to previous approaches. 
\begin{table}[htbp]
\centering
\small
\caption{\textbf{Comparison of different methods on overall-class mAP (\%) using various backbones.} In-domain (id) performance is reported. Rows with gray background indicate our proposed method.}
\label{tab:FPS_FLOPS_Params}
\resizebox{\linewidth}{!}{%
\begin{tabular}{l l r r r r}
\toprule
Method & Backbone & mAP (id) & FLOPS & Trainable Params. & FPS\\
\midrule
MAFT+            & ConvNext-B & 40.08 & 210G  & 108.66M & 12.20\\
OVSeg            & - & 39.26 & 1.84T & 408.55M  & - \\
\rowcolor{gray!10} Our Method (vits) & ConvNext-B & 42.36 & 259G  & 22.12M  & 10.53\\
\rowcolor{gray!10} Our Method (vitb) & ConvNext-B & 44.54 & 365G  & 22.51M  & 9.90\\
\rowcolor{gray!10} Our Method (vitl) & ConvNext-B & 44.37 & 721G  & 22.77M  & 7.52\\
\midrule
MAFT+            & ConvNext-L & 53.41 & 368G  & 223.22M & 9.52\\
OVSeg            & - & 39.26 & 1.84T & 408.55M  & - \\
\rowcolor{gray!10} Our Method (vits) & ConvNext-L & 54.22 & 416G  & 22.33M  & 8.85\\
\rowcolor{gray!10} Our Method (vitb) & ConvNext-L & 55.22 & 522G  & 22.82M  & 8.20\\
\rowcolor{gray!10} Our Method (vitl) & ConvNext-L & 56.71 & 878G  & 23.09M  & 6.49\\
\bottomrule
\end{tabular}%
}
\vspace{-10pt}
\end{table}


\subsection{Qualitative Results.}

\paragraph{Qualitative Performance on Visual-Geometric Fusion.} The qualitative comparisons in \cref{fig:pic_vis_feats} demonstrate that integrating visual and geometric information consistently outperforms using either modality alone.

\paragraph{Qualitative Performance on Segmentation Maps.} In the successful cases (\cref{fig:pic_vis_seg}(a)), we compare our method with other state-of-the-art approaches, namely FCCLIP and EOV-Seg. For diverse underwater organisms like Abalone, Atlantic Spadefish, and Blacktail Butterflyfish, our method demonstrates superior segmentation performance. More qualitative results are in the Appendix I.

\paragraph{Fault Cases Analysis \& Comparison.} As shown in the failure case (\cref{fig:pic_vis_seg}(b)), our model misclassifies \texttt{Anyperodon Leucogrammicus} as \texttt{Peacock Grouper} or \texttt{Coral Hind}, mainly due to their grouper-like morphology with colorful, patterned bodies. This highlights the need for future models to better disentangle visual similarity from semantic distinctiveness.
\section{Conclusion}
\label{sec:conclusion}
We introduced MARIS, the first large-scale fine-grained benchmark for open-vocabulary underwater instance segmentation, addressing the limitations of existing datasets with coarse-grained labels. Our framework integrates \textbf{GPEM} to leverage stable geometric cues and \textbf{SAIM} to enrich language priors, improving segmentation under challenging underwater conditions. Overall, MARIS and the proposed framework provide a robust benchmark and methodology for open-vocabulary segmentation in challenging underwater scenarios.

\textbf{Limitation:} While MARIS covers diverse categories, extreme environments and rare species remain underrepresented, which may limit generalization. Future work will focus on expanding the dataset and enhancing model robustness in such scenarios.

\paragraph{Acknowledgments}
This work was supported in part by the National Natural Science Foundation of China under Grants 62306241 and U62576284.

{
    \small
    \bibliographystyle{ieeenat_fullname}
    \bibliography{main}
}

\clearpage
\appendix
\setcounter{section}{0}
\setcounter{page}{1}
\maketitlesupplementary

\section{Robustness Analysis of SAIM}
\begin{table}[ht]
\centering
\small
\caption{\textbf{Ablation study on TopN selection.} We report mAP, AP$_{50}$, and AP$_{75}$ for different TopN values.}
\label{tab:topn_ablation}
\begin{tabular}{cccc}
\toprule
TopN & mAP & AP$_{50}$ & AP$_{75}$ \\
\midrule
1  & 41.73 & 48.04 & 45.28 \\
2  & 43.83 & 50.45 & 47.61 \\
5  & 43.97 & 50.62 & 47.77 \\
10 & \textbf{44.37} & 50.99 & 48.11 \\
20 & \textbf{44.37} & \textbf{51.41} & 47.90 \\
50 & 44.42 & 51.07 & \textbf{48.18} \\
80 & 44.33 & 51.01 & 48.11 \\
\bottomrule
\end{tabular}
\vspace{-10pt}
\end{table}

The SAIM module demonstrates strong robustness to the choice of TopN. Its template selection mechanism remains stable across different TopN settings, maintaining consistent segmentation performance. This insensitivity reduces the need for extensive hyperparameter tuning and ensures reliable performance.

\section{Implementation Details}

\begin{itemize}
    \item \textbf{EOVSeg}: We set \texttt{NUM\_STAGE} to 1, and adopted \texttt{ViT-B/16} as an auxiliary encoder. 
    For CLIP pre-trained parameters, we experimented with both \texttt{ConvNeXt-B} and \texttt{ConvNeXt-L}.
    \item \textbf{FCCLIP}: The model was configured with \texttt{TRANSFORMER\_ENC\_LAYERS} = 6 and \texttt{DEC\_LAYERS} = 10, and employed CLIP pre-trained weights from both \texttt{ConvNeXt-B} and \texttt{ConvNeXt-L}.
    \item \textbf{MAFT+}: We adopted the same transformer settings (\texttt{TRANSFORMER\_ENC\_LAYERS} = 6 and \texttt{DEC\_LAYERS} = 10), with CLIP pre-training based on \texttt{ConvNeXt-B} and \texttt{ConvNeXt-L}.
    \item \textbf{MARIS}: We followed the same setting as FCCLIP and MAFT+, i.e., \texttt{TRANSFORMER\_ENC\_LAYERS} = 6 and \texttt{DEC\_LAYERS} = 10, with CLIP pre-trained parameters from \texttt{ConvNeXt-B}and \texttt{ConvNeXt-L}.
\end{itemize}

For all other hyperparameters, we followed the original papers.

\section{Code Release:} 
Full code and model weights are available at appendix. 
Includes: 
(1) Preprocessing scripts for MARIS dataset; 
(2) How to intall the environment to start the expriments.
(3) How to run the code to reproduce our results.

\section{Template Selection Strategy}
\paragraph{I. Mixed-based Selection.}  
Given the similarity tensor $\mathcal{S} \in \mathbb{R}^{B \times H \times W \times K \times T}$ between image patches and text templates, we compute the average score across spatial positions:
\begin{equation}
    \bar{\mathcal{S}}_{b,k,t} = \frac{1}{H \cdot W} \sum_{h=1}^{H}\sum_{w=1}^{W} \mathcal{S}_{b,h,w,k,t}, \quad
    \bar{\mathcal{S}} \in \mathbb{R}^{B \times K \times T}.
\end{equation}
For each category $k$, we rank the template indices $t$ according to $\bar{\mathcal{S}}_{b,k,t}$ and select the top-$N$ candidates.  
The corresponding embeddings are gathered and averaged across batches:
\begin{equation}
    \mathbf{E}_{k}^{\text{top}} = \frac{1}{B \cdot N} \sum_{b=1}^{B} \sum_{t \in \text{TopN}(\bar{\mathcal{S}}_{b,k,:})} \mathbf{E}_{k,t}.
\end{equation}
To balance global and local information, the final category embedding is obtained by interpolating between the aggregated top-$N$ features and the overall average embedding:
\begin{equation}
    \mathbf{E}_k = \lambda \cdot \mathbf{E}_{k}^{\text{top}} + (1-\lambda) \cdot \frac{1}{T}\sum_{t=1}^{T}\mathbf{E}_{k,t},
\end{equation}
where $\lambda$ controls the contribution of top-ranked templates. This strategy emphasizes the most discriminative templates while retaining global semantic consistency.

\paragraph{II. Weighted Top-$N$ Enhancement.}  
Alternatively, we introduce an adaptive weighting scheme to explicitly enhance the contribution of high-confidence templates.  
Based on the mean similarity $\bar{\mathcal{S}}_{b,k,t}$, we identify the top-$N$ templates per category $k$ and construct a binary mask $\mathcal{M}_{b,k,t}$ where $\mathcal{M}_{b,k,t}=1$ if $t$ is in the top-$N$ set and $0$ otherwise.  
Each selected template is assigned an enhancement factor $\alpha>1$:
\begin{equation}
    W_{b,k,t} = 
    \begin{cases}
        \alpha, & \text{if } \mathcal{M}_{b,k,t}=1, \\
        1, & \text{otherwise}.
    \end{cases}
\end{equation}
The weights are normalized across templates to form a probability distribution:
\begin{equation}
    \tilde{W}_{b,k,t} = \frac{W_{b,k,t}}{\sum_{t=1}^{T} W_{b,k,t}}.
\end{equation}
The final category embedding is then computed as the weighted sum of template features:
\begin{equation}
    \mathbf{E}_k = \frac{1}{B}\sum_{b=1}^{B}\sum_{t=1}^{T} \tilde{W}_{b,k,t} \cdot \mathbf{E}_{k,t}.
\end{equation}
This strategy adaptively emphasizes high-confidence templates without discarding others, leading to a more robust and discriminative representation.

\paragraph{Practical Consideration.}  
To ensure efficient training and evaluation, we adopt a simplified yet effective strategy by performing template selection with only a single randomly sampled image per category. Although this reduces the computational cost substantially, our experiments demonstrate that even a single image provides sufficient discriminative signal to reliably identify informative templates.

\section{Dataset Diversity Analysis}
\label{app:dataset_diversity}

\begin{figure*}[htbp]
    \centering
    \includegraphics[width=0.75\linewidth]{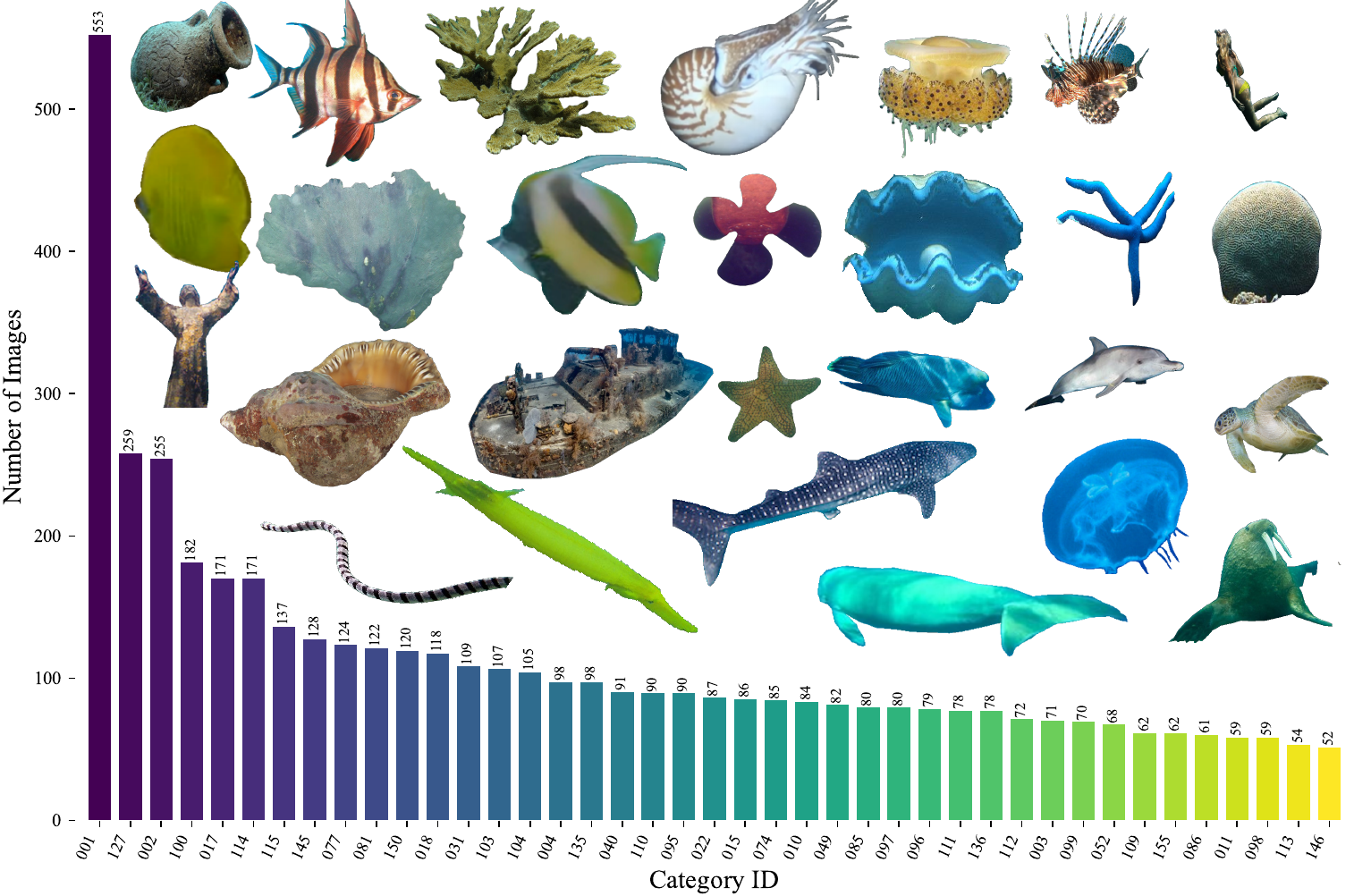}
    \caption{\textbf{Instance distribution of Intersection Classes in MARIS validation set.} 
    Shows the number of instances for classes shared between training and validation sets.}
    \label{fig:intersection_class_counts}
\end{figure*}

\begin{figure*}[htbp]
    \centering
    \includegraphics[width=0.75\linewidth]{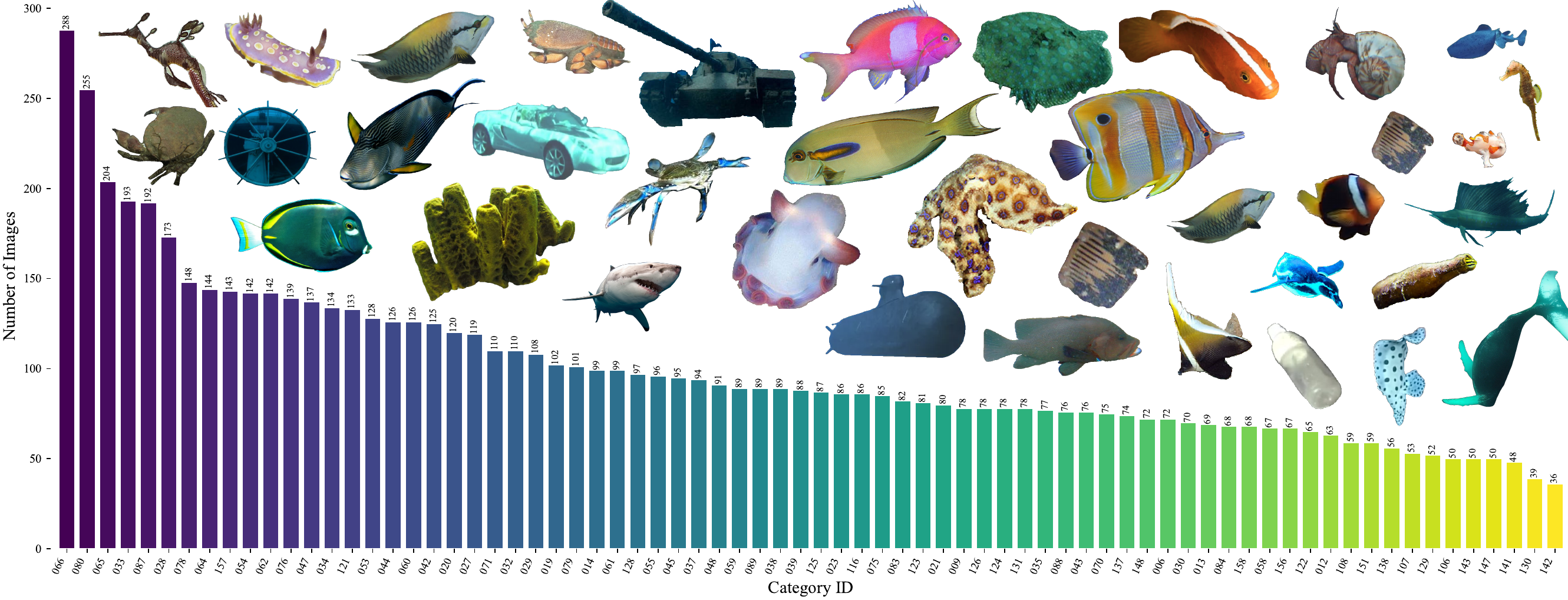}
    \caption{\textbf{Instance distribution of Open-Vocabulary (OV) Classes in MARIS validation set.} 
    Shows the number of instances for classes that appear only in validation.}
    \label{fig:ov_class_counts}
\end{figure*}

\begin{figure*}[htbp]
    \centering
    \includegraphics[width=\linewidth]{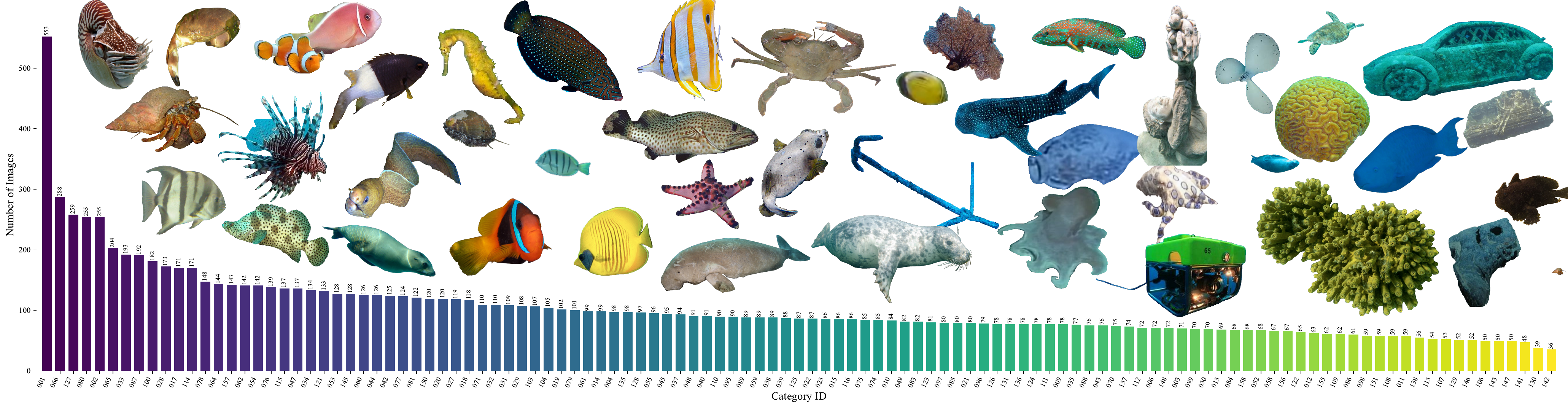}
    \caption{\textbf{Instance distribution of Overall Classes in MARIS validation set.} 
    Provides the counts for all classes, giving an overall view of dataset composition and class imbalance.}
    \label{fig:overall_class_counts}
\end{figure*}

\paragraph{Instance Diversity.} To provide a comprehensive understanding of category coverage in MARIS, we analyze the distribution of instances across the validation set, as illustrated in \cref{fig:intersection_class_counts}-\cref{fig:overall_class_counts}. We visualize the relationship between instance counts and category IDs across different splits of the MARIS dataset.  
\cref{fig:intersection_class_counts} reports the distribution of intersection classes shared between training and validation, revealing substantial imbalance where frequent species (e.g., common reef fish) dominate the samples, while rare species contain fewer than 60 instances.  
\cref{fig:ov_class_counts} focuses on the open-vocabulary (OV) classes that appear only in the validation set. Although MARIS contains 74 OV categories, their frequency varies significantly, indicating that models must handle long-tailed distributions when generalizing to unseen classes.  
Finally, \cref{fig:overall_class_counts} presents the overall class distribution, highlighting the combined imbalance across both seen and unseen categories.  

This analysis demonstrates that MARIS is not only fine-grained but also diverse, covering a wide range of marine organisms, man-made objects, and substrates. At the same time, the inherent long-tailed distribution reflects real-world underwater environments, where rare species often occur sparsely. Thus, MARIS provides a challenging yet realistic benchmark for evaluating the generalization ability of open-vocabulary segmentation models.

\paragraph{Category Diversity.} 
Following the parent category taxonomy defined in \cite{hong2025usis16k}, we analyze the category diversity of our dataset, as summarized in Tables~\ref{tab:category_Train}, \ref{tab:category_Only_In_Train}, and \ref{tab:combined_category_validation}. This analysis highlights the extensive coverage of both common and rare underwater object classes, illustrating the richness of our dataset. Compared to previous datasets such as WaterMask \cite{lian2023watermask} and UWSAM \cite{li2025uwsam}, our dataset not only includes a broader set of categories but also demonstrates a more balanced and rational parent category organization. The breakdown into Intersection, OV, and Overall classes further supports the validity of our category design, emphasizing the dataset's potential for training robust models and evaluating generalization across diverse underwater scenarios.

\section{Dataset Image Feature Analysis}
\label{sec:underwater_validation_analysis}
\begin{figure*}[htbp]
    \centering
    \includegraphics[width=\linewidth]{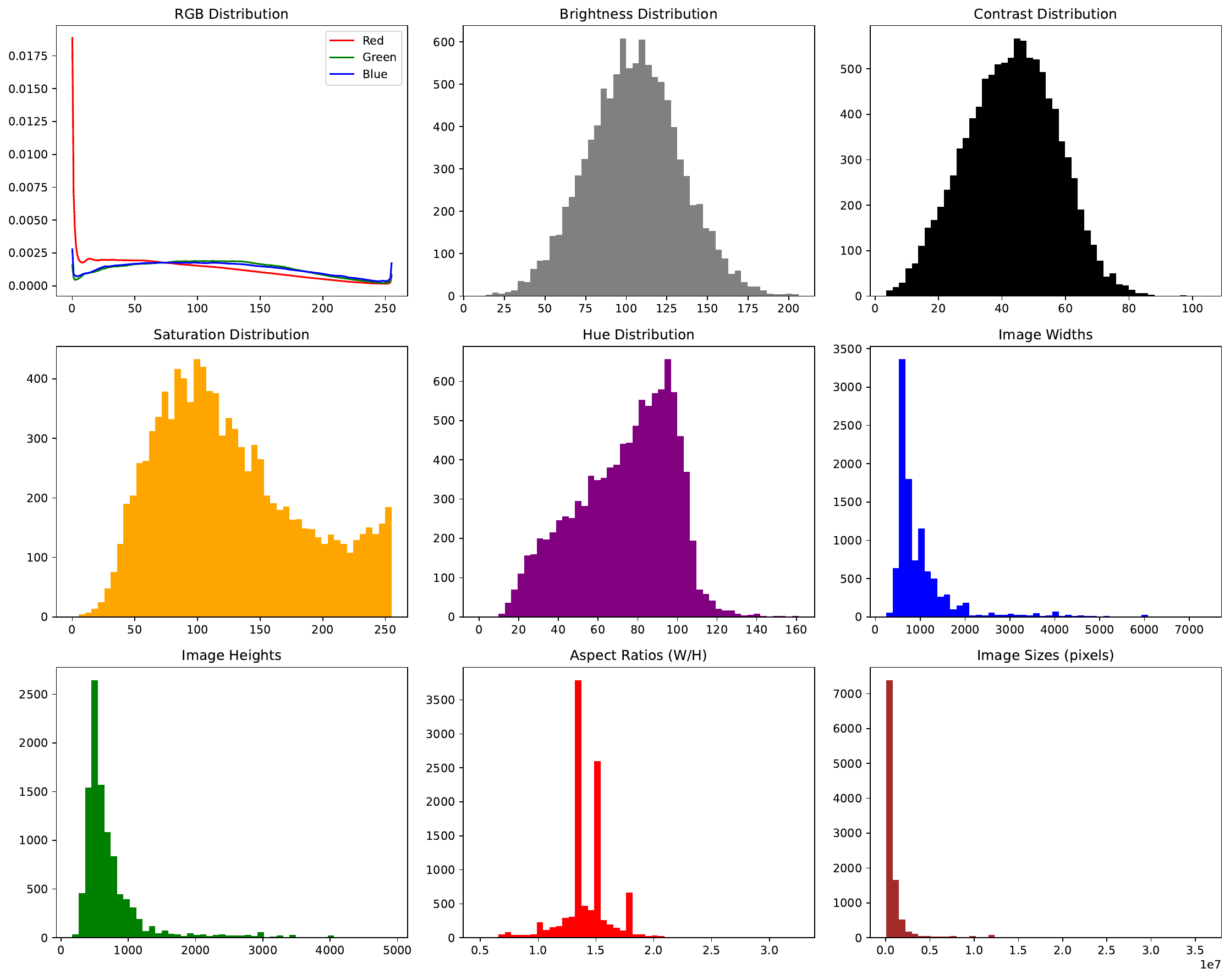}
    \caption{\textbf{Validation Set Image Feature Analysis.} 
    Comprehensive analysis of the underwater validation set across nine dimensions, including \emph{color space} (RGB distribution, hue, saturation), \emph{perceptual quality} (contrast, brightness), and \emph{geometric attributes} (width, height, resolution, aspect ratio).}
    \label{fig:sup_dataset_analysis}
\end{figure*}

The underwater validation set is analyzed across nine dimensions (in \cref{fig:sup_dataset_analysis}), spanning color space, perceptual quality, and geometric attributes. These distributions reveal characteristics highly adapted to underwater imaging conditions, providing crucial support for model evaluation in this domain.  
\textbf{Color space.} The RGB channels exhibit balanced distributions within the 0--250 intensity range, with frequencies concentrated in mid-level values (300--500 counts), mitigating bias from single-color dominance caused by light scattering. Hue follows a ``middle-high, low-at-extremes'' distribution with peaks around 400 counts, reflecting the prevalence of neutral tones consistent with water transparency and plankton density. Saturation is concentrated in the 40--120 range (500--600 counts), with low contributions at both extremes, aligning with the natural attenuation of vivid colors caused by underwater light refraction.  
\textbf{Perceptual quality.} Contrast shows a monotonic increase across the 0--100 range, peaking at 600 counts within 80--100, which counteracts blurring induced by turbidity. Brightness values are concentrated in the 100--200 range with probability density 0.015--0.0175, corresponding well to illumination variations across depths, thus ensuring visual clarity and feature discriminability.  
\textbf{Geometric attributes.} Image width (0--7000 pixels) and height (0--5000 pixels) are concentrated in mid-scales, with peaks in 2000--4000 (width, 3500 counts) and 2000--3000 (height, 2500 counts). Image sizes in the 2$\times10^6$--6$\times10^6$ pixel range dominate (7000 counts). Aspect ratios are primarily distributed between 1.0--2.0 (peak 3500 counts), which matches standard underwater camera formats while preserving object integrity for targets such as corals and fish.  
Overall, the validation set exhibits feature distributions that align closely with underwater optical characteristics, environmental conditions, and imaging requirements, thereby providing a reliable basis for assessing model generalization in tasks such as underwater object detection and scene segmentation.  

\section{Acknowledgement of Data Sources}
We would like to formally acknowledge the contributions of the following datasets, which serve as the foundation for MARIS. The WaterMask \cite{lian2023watermask} dataset provides richly annotated underwater imagery for diverse scene understanding tasks. Additionally, the recently released underwater datasets USIS16K \cite{hong2025usis16k}, UWSAM \cite{li2025uwsam}, and the semantic segmentation dataset by \cite{islam2020semantic} have been systematically re-annotated and extended to ensure consistency and comprehensive coverage. We are grateful for the efforts of the original dataset creators, whose careful data collection and annotation make this work possible.

\section{Underwater Prompts}
\label{app:uw_prompts}
To effectively adapt text embeddings to underwater semantics, we design a comprehensive collection of domain-aware prompt templates. Beyond generic templates (e.g., ``a photo of a \{\}''), our design incorporates five additional dimensions that explicitly capture the unique characteristics of underwater imagery: \emph{environment}, \emph{medium/visibility}, \emph{lighting}, \emph{depth}, and \emph{scene interaction}, as summarized in Appendix \cref{app:prompt_1}-\cref{app:prompt_3}.  

Environment-oriented prompts describe contextual backgrounds such as coral reefs, caves, or shipwrecks (e.g., ``a \{\} near a coral reef''), which provide strong location priors. Medium/visibility prompts reflect variations in water clarity, ranging from crystal-clear tropical seas to turbid or plankton-rich conditions (e.g., ``a \{\} in low visibility conditions''), thereby modeling visual degradations that frequently occur underwater. Lighting prompts capture distinct illumination conditions including bioluminescence, diver flashlights, or strong sunlight filtering through the water column (e.g., ``a \{\} illuminated by artificial light underwater''), which are crucial for robust representation learning under diverse visual appearances. Depth-related prompts explicitly encode the ecological and physical differences across ocean layers, from shallow reefs to the hadal trenches (e.g., ``a \{\} at mesopelagic depth''), helping the model disambiguate species that are depth-specific. Finally, scene/interaction prompts describe dynamic relationships such as co-occurrence, interactions with divers or vehicles, and natural behaviors (e.g., ``a \{\} swimming with other fish underwater''), which improve context awareness.  

By enriching textual representations with these underwater-specific prompts, our method bridges the semantic gap between terrestrial-pretrained vision–language models and the marine domain. Empirical results in \cref{tab:ablation_eg_ev} confirm that the combination of prompt engineering and adaptive template selection consistently improves both overall segmentation accuracy and open-vocabulary generalization, demonstrating the importance of underwater-aware textual priors in guiding vision–language alignment.

\begin{table*}[ht]
\centering
\resizebox{\textwidth}{!}{
\begin{tabular}{lp{0.8\textwidth}}
\hline
\textbf{Parent Category} & \textbf{Child Category (Train)} \\
\hline
Human & Diver, Swimmer \\
\hline
Fish & Achilles Tang, Anampses Twistii, Bicolor Angelfish, Blue Parrotfish, Blue-spotted Wrasse, Bluecheek Butterflyfish, Bullhead Shark, Enoplosus Armatus, Giant Wrasse, Graysby, Hammerhead Shark, Lined Surgeonfish, Lionfish, Manta Ray, Mirror Butterflyfish, Mola, Moorish Idol, Moray Eel, Orbicular Batfish, Potato Grouper, Redsea Bannerfish, Regal Blue Tang, Saddle Butterflyfish, Sawfish, Spotted Wrasse, Stoplight Parrotfish, Threadfin Butterflyfish, Trumpetfish, Twin-spot Goby, Whale Shark, Whitespotted Surgeonfish \\
\hline
Non fish & Brain Coral, Common Octopus, Common Prawn, Crinoid, Dolphin, Dugong, Elkhorn Coral, Fan Coral, Fried Egg Jellyfish, Geoduck, Giant Clams, Killer Whale, King Crab, Linckia Laevigata, Lion's Mane Jellyfish, Manatee, Mantis Shrimp, Moon Jellyfish, Nautilus, Oreaster Reticulatus, Protoreaster Nodosus, Scallop, Sea Anemone, Sea Cucumber, Sea Lion, Sea Urchin, Snake, Spiny Lobster, Squid, Triton's Trumpet, Turtle, Walrus \\
\hline
Marine Garbage & Can, Plastic Bag, Surgical Mask, Tyre \\
\hline
Wrecked Vehicle & Shipwreck, Wrecked Aircraft \\
\hline
Lost item & Gun, Phone \\
\hline
Archeology & Amphora, Coin, Statue \\
\hline
Underwater equipment & Autonomous Underwater Vehicle (AUV), Personal Submarines, Remotely Operated Vehicle (ROV) \\
\hline
Underwater operation & Over Board Valve, Propeller, Ship's Anode \\
\hline
\end{tabular}}
\caption{\textbf{Category Diversity Analysis for Train dataset.} This table presents a detailed breakdown of parent categories in the dataset, highlighting the diversity of objects in the training set.}
\label{tab:category_Train}
\end{table*}

\begin{table*}[ht]
\centering
\resizebox{\textwidth}{!}{
\begin{tabular}{lp{0.8\textwidth}}
\hline
\textbf{Parent Category} & \textbf{Child Category(Only in Train)} \\
\hline
Human &  \\
\hline
Fish & Achilles Tang, Anampses Twistii, Bicolor Angelfish, Bullhead Shark, Graysby, Lined Surgeonfish, Manta Ray, Mirror Butterflyfish, Mola, Moorish Idol, Orbicular Batfish, Potato Grouper, Regal Blue Tang, Saddle Butterflyfish, Sawfish, Spotted Wrasse, Stoplight Parrotfish, Twin-spot Goby, Whitespotted Surgeonfish \\
\hline
Non fish & Common Octopus, Common Prawn, Crinoid, Killer Whale, King Crab, Lion's Mane Jellyfish, Mantis Shrimp, Scallop, Sea Anemone, Sea Cucumber, Spiny Lobster, Squid \\
\hline
Marine Garbage & Can, Surgical Mask, Tyre \\
\hline
Wrecked Vehicle &  \\
\hline
Lost item & Gun, Phone \\
\hline
Archeology & Coin \\
\hline
Underwater equipment & Autonomous Underwater Vehicle (AUV), Personal Submarines \\
\hline
Underwater operation & Over Board Valve, Ship's Anode \\
\hline
\end{tabular}}
\caption{\textbf{Category Diversity Analysis for Class Only in Train dataset.} This table presents a detailed breakdown of parent categories in the dataset, highlighting the diversity of objects in the training set.}
\label{tab:category_Only_In_Train}
\end{table*}

\begin{table*}[ht]
\centering
\small
\resizebox{\textwidth}{!}{
\begin{tabular}{>{\centering\arraybackslash}m{3cm}p{0.25\textwidth}p{0.25\textwidth}p{0.35\textwidth}}
\hline
\textbf{Parent Category} & \textbf{Child Category (Intersection)} & \textbf{Child Category (OV)} & \textbf{Child Category (Overall)} \\
\hline
Human & Diver, Swimmer &  & Diver, Swimmer \\
\hline
Fish & Blue Parrotfish, Blue-spotted Wrasse, Bluecheek Butterflyfish, Enoplosus Armatus, Giant Wrasse, Hammerhead Shark, Lionfish, Moray Eel, Redsea Bannerfish, Threadfin Butterflyfish, Trumpetfish, Whale Shark 
& Anyperodon Leucogrammicus, Atlantic Spadefish, Blackspotted Puffer, Blacktail Butterflyfish, Chromis Dimidiata, Cinnamon Clownfish, Convict Surgeonfish, Copperband Butterflyfish, Coral Hind, Electric Ray, Eritrean Butterflyfish, Fire Goby, Flounder, Frogfish, Great White Shark, Heniochus Varius, Hippocampus, Humpback Grouper, Lunar Fusilier, Maldives Damselfish, Ocellaris Clownfish, Orange Skunk Clownfish, Orange-band Surgeonfish, Peacock Grouper, Pink Anemonefish, Pomacentrus Sulfureus, Porcupinefish, Porkfish, Powder Blue Tang, Pseudanthias Pleurotaenia, Pyramid Butterflyfish, Raccoon Butterflyfish, Red-breasted Wrasse, Redmouth Grouper, Sailfish, Scissortail Sergeant, Sea Dragon, Slingjaw Wrasse, Sohal Surgeonfish, Spotted Drum, Threespot Angelfish, Thresher Shark, Whitecheek Surgeonfish, Yellow Boxfish
& Anyperodon Leucogrammicus, Atlantic Spadefish, Blackspotted Puffer, Blacktail Butterflyfish, Blue Parrotfish, Blue-spotted Wrasse, Bluecheek Butterflyfish, Chromis Dimidiata, Cinnamon Clownfish, Convict Surgeonfish, Copperband Butterflyfish, Coral Hind, Electric Ray, Enoplosus Armatus, Eritrean Butterflyfish, Fire Goby, Flounder, Frogfish, Giant Wrasse, Great White Shark, Hammerhead Shark, Heniochus Varius, Hippocampus, Humpback Grouper, Lionfish, Lunar Fusilier, Maldives Damselfish, Moray Eel, Ocellaris Clownfish, Orange Skunk Clownfish, Orange-band Surgeonfish, Peacock Grouper, Pink Anemonefish, Pomacentrus Sulfureus, Porcupinefish, Porkfish, Powder Blue Tang, Pseudanthias Pleurotaenia, Pyramid Butterflyfish, Raccoon Butterflyfish, Red-breasted Wrasse, Redmouth Grouper, Redsea Bannerfish, Sailfish, Scissortail Sergeant, Sea Dragon, Slingjaw Wrasse, Sohal Surgeonfish, Spotted Drum, Threadfin Butterflyfish, Threespot Angelfish, Thresher Shark, Trumpetfish, Whale Shark, Whitecheek Surgeonfish, Yellow Boxfish \\
\hline
Non fish & Brain Coral, Dolphin, Dugong, Elkhorn Coral, Fan Coral, Fried Egg Jellyfish, Geoduck, Giant Clams, Linckia Laevigata, Manatee, Moon Jellyfish, Nautilus, Oreaster Reticulatus, Protoreaster Nodosus, Sea Lion, Sea Urchin, Snake, Triton's Trumpet, Turtle, Walrus
& Abalone, Blue-ringed Octopus, Cancer Pagurus, Dumbo Octopus, Hermit Crab, Homarus, Humpback Whale, Penguin, Queen Conch, Sea Slug, Seal, Spanner Crab, Sperm Whale, Sponge, Swimming Crab
& Abalone, Blue-ringed Octopus, Brain Coral, Cancer Pagurus, Dolphin, Dugong, Dumbo Octopus, Elkhorn Coral, Fan Coral, Fried Egg Jellyfish, Geoduck, Giant Clams, Hermit Crab, Homarus, Humpback Whale, Linckia Laevigata, Manatee, Moon Jellyfish, Nautilus, Oreaster Reticulatus, Penguin, Protoreaster Nodosus, Queen Conch, Sea Lion, Sea Slug, Sea Urchin, Seal, Snake, Spanner Crab, Sperm Whale, Sponge, Swimming Crab, Triton's Trumpet, Turtle, Walrus \\
\hline
Marine Garbage & Plastic Bag & Glass Bottle, Plastic Bottle, Plastic Box, Plastic Cup & Glass Bottle, Plastic Bag, Plastic Bottle, Plastic Box, Plastic Cup \\
\hline
Wrecked Vehicle & Shipwreck, Wrecked Aircraft & Wrecked Car, Wrecked Tank & Shipwreck, Wrecked Aircraft, Wrecked Car, Wrecked Tank \\
\hline
Lost item &  & Boots, Glasses, Ring & Boots, Glasses, Ring \\
\hline
Archeology & Amphora, Statue & Anchor, Ship's Wheel & Amphora, Anchor, Ship's Wheel, Statue \\
\hline
Underwater equipment & Remotely Operated Vehicle (ROV) & Military Submarines & Military Submarines, Remotely Operated Vehicle (ROV) \\
\hline
Underwater operation & Propeller & Pipeline's Anode, Sea Chest Grating, Submarine Pipeline & Pipeline's Anode, Propeller, Sea Chest Grating, Submarine Pipeline \\
\hline
\end{tabular}}
\caption{\textbf{Combined Category Diversity for Validation Dataset.} This table integrates Intersection Class, OV Class, Overall Class for each parent category. It provides a comprehensive overview of category coverage and diversity, highlighting both shared and unique classes.}
\label{tab:combined_category_validation}
\end{table*}

\begin{table*}[ht]
\centering
\renewcommand{\arraystretch}{1.2}
\setlength{\tabcolsep}{6pt}
\begin{tabular}{p{0.47\linewidth} p{0.47\linewidth}}
\toprule
\textbf{Generic Prompt} & \textbf{Environment / Background} \\
\midrule
a photo of a \{\} & a \{\} underwater \\
This is a photo of a \{\} & a \{\} in the ocean \\
There is a \{\} in the underwater scene & a \{\} in the deep sea \\
a photo of a \{\} in \{\} & a \{\} near a coral reef \\
a photo of a small \{\} & a \{\} in murky underwater conditions \\
a photo of a medium \{\} & a \{\} in a tropical sea \\
a photo of a large \{\} & a \{\} in a freshwater lake \\
This is a photo of a small \{\} & a \{\} in brackish water \\
This is a photo of a medium \{\} & a \{\} in shallow coastal water \\
This is a photo of a large \{\} & a \{\} in open ocean water \\
\bottomrule
\end{tabular}
\caption{Prompt templates for \textbf{Generic} and \textbf{Environment/Background} categories.}
\label{app:prompt_1}
\end{table*}

\begin{table*}[ht]
\centering
\renewcommand{\arraystretch}{1.2}
\setlength{\tabcolsep}{6pt}
\begin{tabular}{p{0.47\linewidth} p{0.47\linewidth}}
\toprule
\textbf{Medium / Visibility} & \textbf{Lighting / Visual} \\
\midrule
a \{\} in turbid blue-green water & a \{\} illuminated by artificial light underwater \\
a \{\} in crystal-clear water & a \{\} glowing in bioluminescent light \\
a \{\} in highly murky water & a \{\} under dim moonlight underwater \\
a \{\} in hazy underwater environment & a \{\} highlighted by a diver’s flashlight \\
a \{\} in water filled with plankton & a \{\} glowing faintly in darkness \\
a \{\} in low visibility conditions & a \{\} in high-contrast underwater light \\
a \{\} in silted water & a \{\} in strong sunlight filtering from above \\
a \{\} in cloudy water & a \{\} in shimmering caustics underwater \\
a \{\} in algae-rich water & a \{\} under soft ambient blue light \\
a \{\} in dark underwater conditions & a \{\} in backlit silhouette underwater \\
\bottomrule
\end{tabular}
\caption{Prompt templates for \textbf{Medium/Visibility} and \textbf{Lighting/Visual} categories.}
\label{app:prompt_2}
\end{table*}

\begin{table*}[ht]
\centering
\renewcommand{\arraystretch}{1.2}
\setlength{\tabcolsep}{6pt}
\begin{tabular}{p{0.47\linewidth} p{0.47\linewidth}}
\toprule
\textbf{Depth / Distance} & \textbf{Scene / Interaction} \\
\midrule
a \{\} at shallow depth near surface & a \{\} surrounded by bubbles \\
a \{\} at mesopelagic depth & a \{\} swimming with other fish underwater \\
a \{\} at bathypelagic depth & a \{\} near a diver underwater \\
a \{\} in the hadal zone trench & a \{\} next to an underwater vehicle \\
close-up of the \{\} underwater & a \{\} entangled in fishing net underwater \\
a \{\} seen from a distance underwater & a \{\} resting near coral \\
a \{\} disappearing into darkness & a \{\} hiding under rocks \\
a \{\} approaching the camera underwater & a \{\} camouflaged in sand \\
a \{\} drifting into the distance & a \{\} gliding through seaweed \\
a \{\} hovering at seabed depth & a \{\} chasing prey underwater \\
\bottomrule
\end{tabular}
\caption{Prompt templates for \textbf{Depth/Distance} and \textbf{Scene/Interaction} categories.}
\label{app:prompt_3}
\end{table*}


\section{More Qualitative Results.}
\begin{figure}[htbp]
    \centering
    \includegraphics[width=0.8\linewidth]{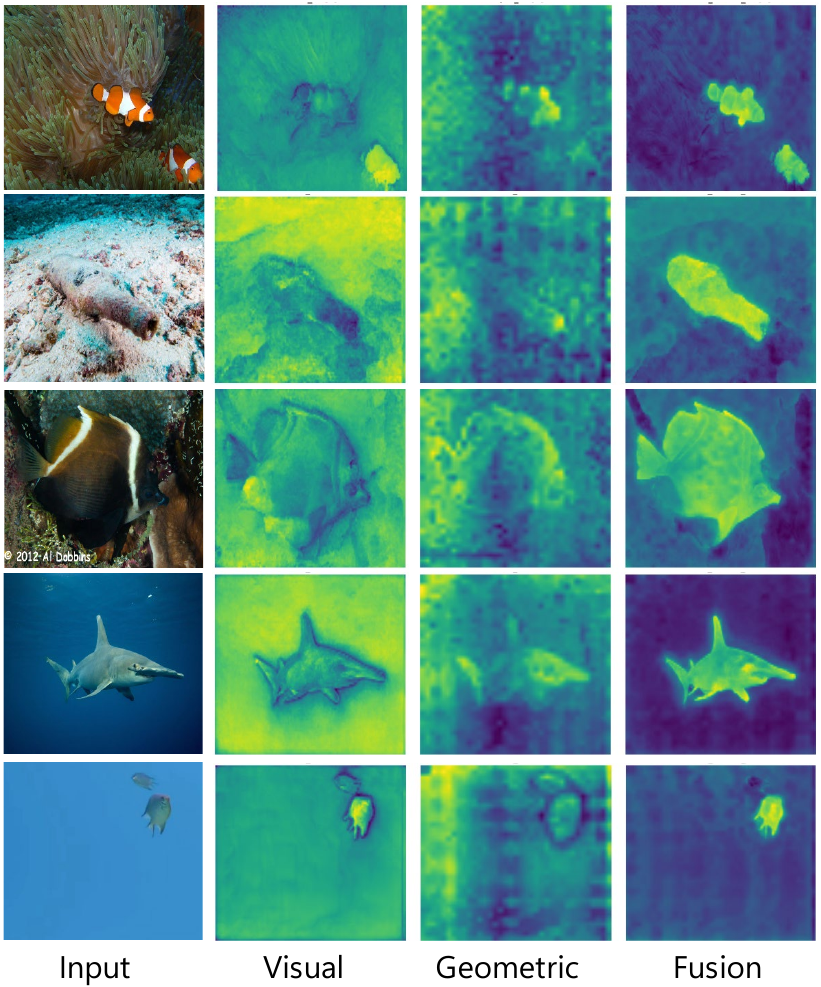}
    \caption{\textbf{Additional Qualitative Results on geometric-enhanced fusion features}}
    \label{fig:sup_vis_feats_vir}
\end{figure}

\begin{figure}[htbp]
    \centering
    \includegraphics[width=0.8\linewidth]{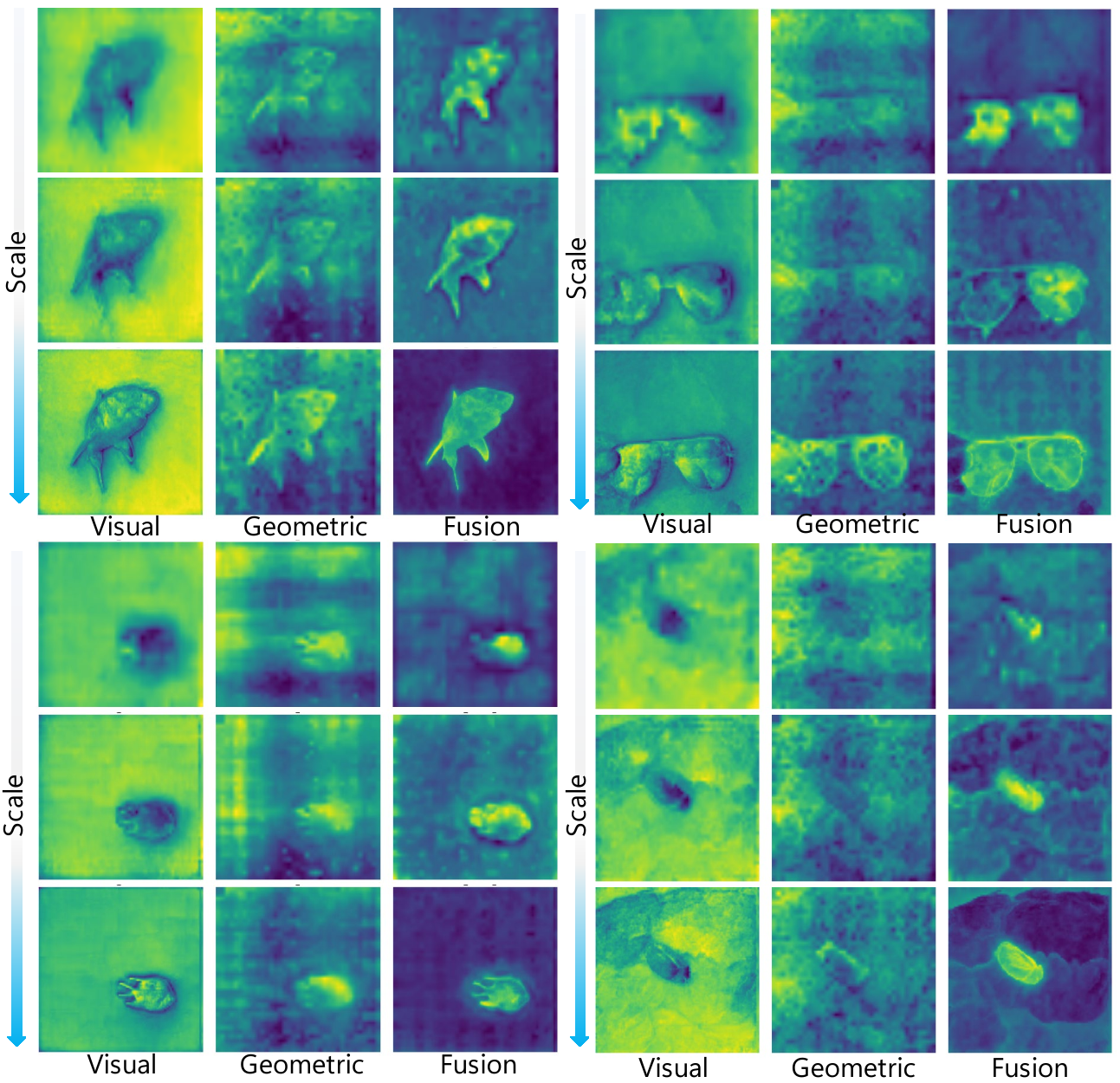}
    \caption{\textbf{Additional Qualitative Results on geometric-enhanced fusion features}}
    \label{fig:sup_vis_feats_vir2}
\end{figure}

\begin{figure}[htbp]
    \centering
    \includegraphics[width=0.8\linewidth]{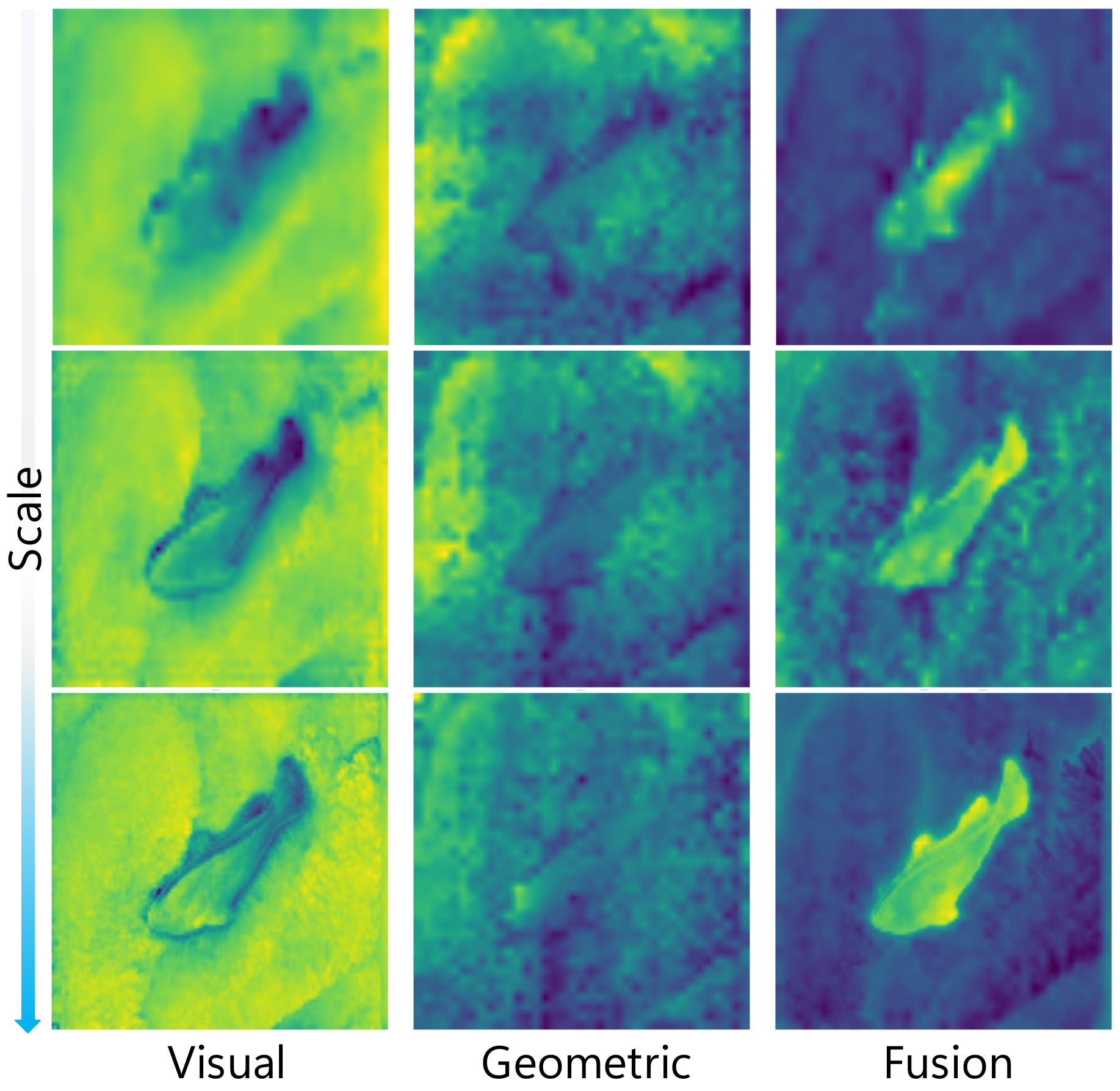}
    \caption{\textbf{Additional Qualitative Results on geometric-enhanced fusion features}}
    \label{fig:sup_vis_feats_vir3}
\end{figure}

\begin{figure}[htbp]
    \centering
    \includegraphics[width=0.85\linewidth]{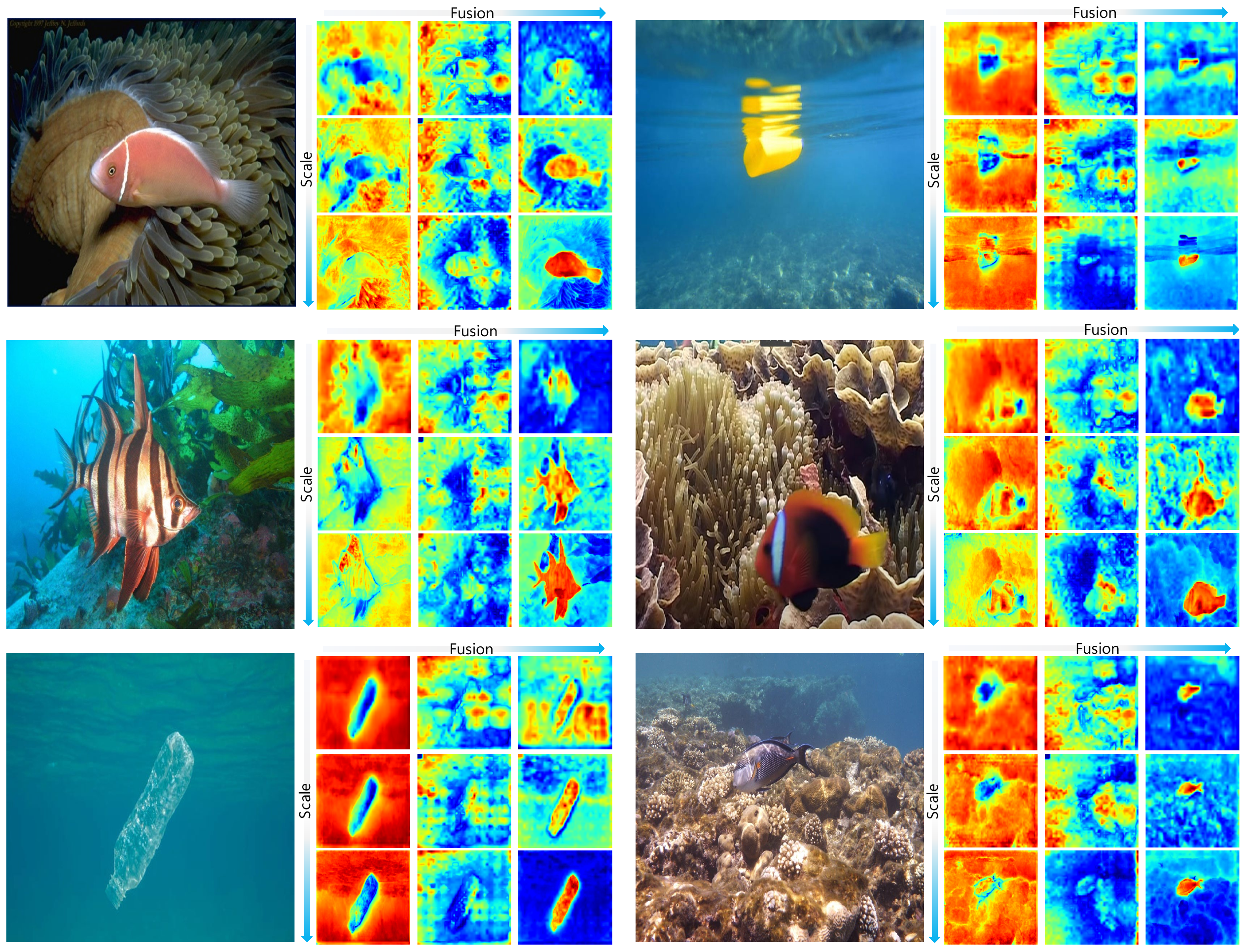}
    \caption{\textbf{Additional Qualitative Results on geometric-enhanced fusion features}}
    \label{fig:sup_vis_feats_jet}
\end{figure}

We present additional qualitative and visualization results (in \cref{fig:sup_vis_feats_vir} - \cref{fig:sup_vis_feats_jet}), where the internal feature visualizations further support the effectiveness of our proposed method. The final segmentation map comparisons demonstrate improved model confidence and enhanced prediction capability.


\section{More Per-Class Experiment Results.}
\begin{figure*}[htbp]
    \centering
    \includegraphics[width=\linewidth]{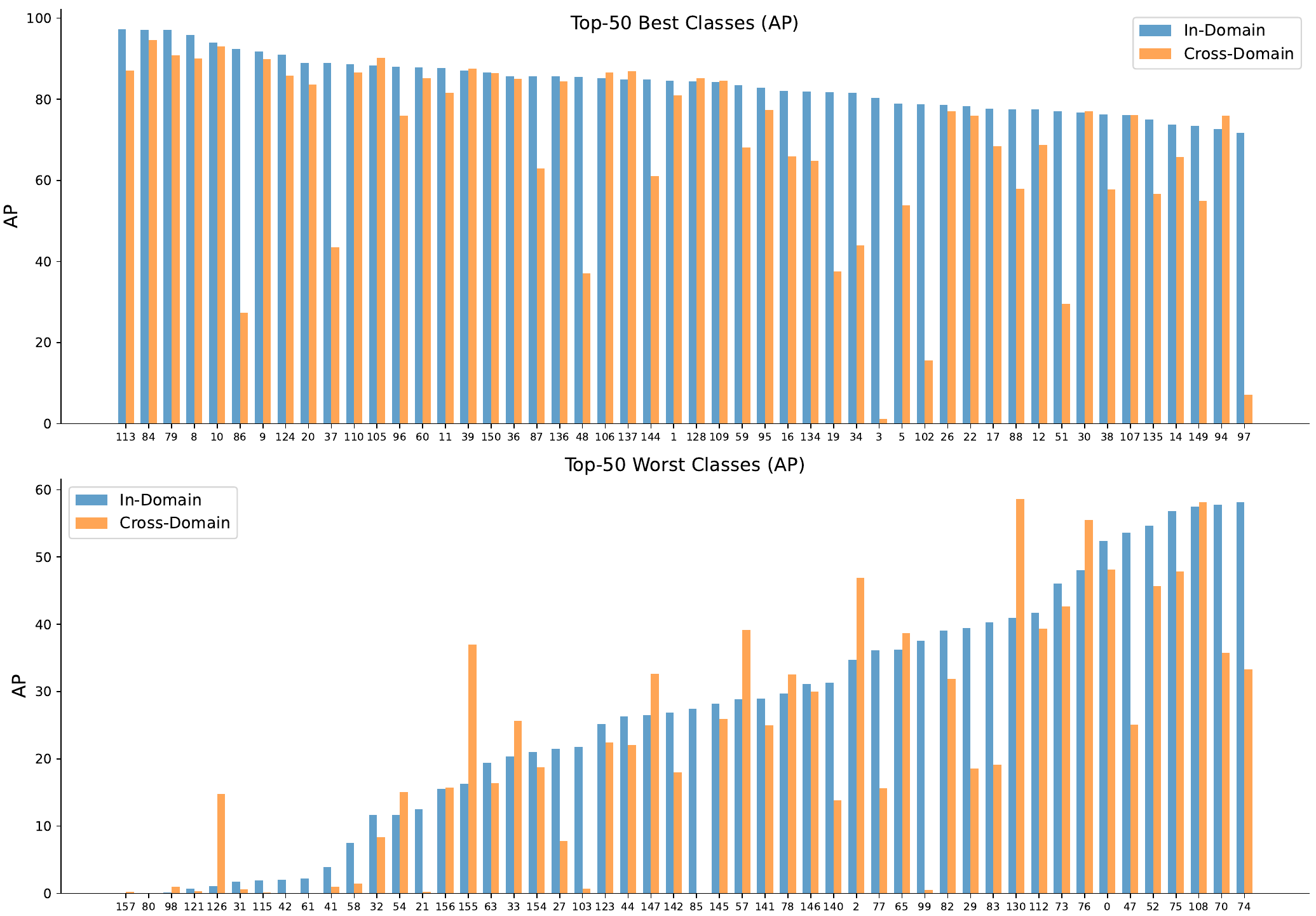}
    \caption{\textbf{Per-class performance comparison under In-Domain and Cross-Domain settings.} 
    Shows how domain shifts affect AP across different classes.}
    \label{fig:sup_top50_per_class}
\end{figure*}

We further present the per-class performance in \cref{fig:sup_top50_per_class}, using category IDs on the x-axis for clearer visualization. We report results for the top-50 best- and worst-performing classes. Consistent with our earlier findings, the In-Domain setting generally outperforms the Cross-Domain setting, highlighting the importance of underwater scene adaptation to improve model performance and suggesting the need for more extensive underwater datasets. Notably, our model achieves superior Cross-Domain performance on certain categories, likely due to the broad coverage of the COCO dataset combined with the strong adaptability of our GPEM and SAIM methods to underwater scenarios.

\end{document}